\documentclass[letterpaper]{article} % DO NOT CHANGE THIS
\usepackage[preprint]{aaai2027}  % DO NOT CHANGE THIS
\usepackage[hyphens]{url}  % DO NOT CHANGE THIS
\usepackage{graphicx} % DO NOT CHANGE THIS
\usepackage{natbib}  % DO NOT CHANGE THIS AND DO NOT ADD ANY OPTIONS TO IT
\usepackage{caption} % DO NOT CHANGE THIS AND DO NOT ADD ANY OPTIONS TO IT
\usepackage{algorithm}
\usepackage{algpseudocode}

\usepackage{newfloat}
\usepackage{listings}
\DeclareCaptionStyle{ruled}{labelfont=normalfont,labelsep=colon,strut=off} % DO NOT CHANGE THIS
\floatstyle{ruled}
\newfloat{listing}{tb}{lst}{}
\floatname{listing}{Listing}

\usepackage{booktabs, multirow}
\usepackage{amsmath, amsfonts, amssymb}
\usepackage{dsfont}
\usepackage{subcaption}
\usepackage{todonotes}
\title{Composing Flow-Matching Energies with Known Physics: Generation, OOD Detection, and Inversion on PDE Fields}
\author {
    Yixuan Sun\textsuperscript{\rm 1,}\thanks{Corresponding author: \texttt{yixuan.sun@anl.gov}},
    Anirban Samaddar\textsuperscript{\rm 1},
    Sandeep Madireddy\textsuperscript{\rm 1}
}
\affiliations {
    \textsuperscript{\rm 1}Mathematics and Computer Science, Argonne National Laboratory\\
  
}

\begin{document}

\maketitle

\begin{abstract}

Probabilistic modeling of physical fields benefits from both a data-driven prior and known physical structure such as the governing equations. Energy-based models (EBMs) are a natural fit since energies compose additively, which enables augmenting physics information during inference. However, EBMs have been difficult to train and sample from due to the intractable partition function. We show in this work that flow matching models with a potential-induced velocity yield an explicit scalar energy at all transport times, whose gradient is exactly the converted learned score and which recovers the marginal negative log-density at the population optimum. The time-dependent energy functions are obtained purely from the matching regression objective on an independent linear Gaussian interpolation, without a variational form or additional MCMC steps, and the sampling retains the flow ODE. Access to the energy function from a trained model serves three roles: energy-corrected data generation, energy as a scoring function for out-of-distribution (OOD) detection, and energy compositional posterior sampling for inverse problems. In particular, we show the explicit energy permits general MCMC samplers in the predictor-corrector sampling framework, reducing PDE residual and spectral distance compared to the flow ODE baseline. Furthermore, we demonstrate utilizing the data energy and physics-based energy (e.g., PDE residuals) as complementary mechanisms to improve detection accuracy for OOD tasks. In addition, we explore the connection to MCMC-based inference for inverse problems by composing the energy with a quadratic observational likelihood that yields a posterior energy, used as an explicitly chosen family of inference-time targets.
% Finally, composing the energy with a quadratic observational likelihood yields a posterior energy that supports MCMC-based inference for inverse problems \AS{Maybe rephrase with highlighting the utility of the approach in inverse problems?}.

\end{abstract}

% Uncomment the following to link to your code, datasets, an extended version or similar.
% You must keep this block between (not within) the abstract and the main body of the paper.
% Make sure that you do not de-anonymize yourself with these links.
% \begin{links}
%     \link{Code}{https://aaai.org/example/code}
%     \link{Datasets}{https://aaai.org/example/datasets}
%     \link{Extended version}{https://aaai.org/example/extended-version}
% \end{links}

  % \textbf{Main message}: energy-based model as read-out from diffusion/flow matching model training, equipped with physics-based prior, useful for OOD, inverse problems (data assimilation with OceanFM), uncertainty quantification, and compositional generation.

  % \textbf{Contribution}: physics prior addressing failure modes of obtained energy from score matching training, resulting energy for generation refinement via composition, enables Boltzmann generator for inverse problems (need to point out what's wrong with diffusion posterior sampling for inverse)

  % \textbf{Related Work} (to compare against)
  % \begin{itemize}
  %   \item EBMs and diffusion/flow models
  %   \item EBMs for OOD and inverse problems (FlowDPS, DiffuisonPDE, ALPS)
  %   \item EBMs for compositional generation (FunFlow, CCFM)
  %   \item EBMs for PDE datasets
  %   \item EBMs failure modes from score matching trianing (EBM classifier work; other
  %         works showing the)
  % \end{itemize}

\section{Introduction}
\begin{figure}
\includegraphics[width=\linewidth]{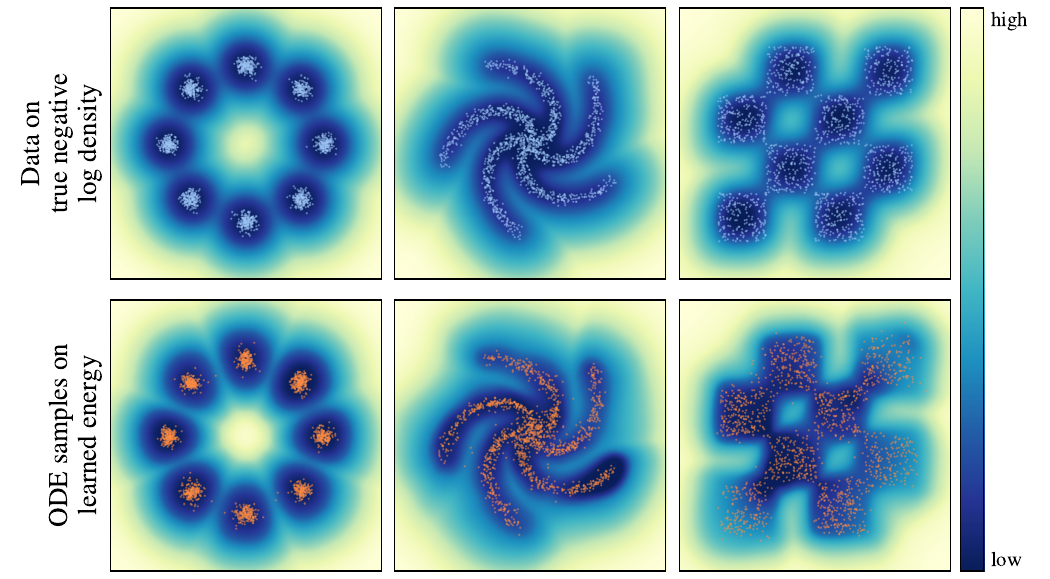}
	\caption{Energy landscapes and flow generated samples of 2D datasets from
	potential parameterized flow matching training. The flow ODE samplers correctly
	generate samples resembling true data points. At the same time, the learned
	energy landscapes share visually the same geometries as the true negative log
	densities, where a lower value represents higher likelihood. }\label{fig:demo}
\vspace{-1em}
\end{figure}
%\ys{should also motivate why the three tasks are common and important in scientific applications}

Flow matching models \cite{Lipman2022-fd, Tong2023-yf} have been widely used for
high-dimensional data generation, including natural images as well as scientific
data. In scientific machine learning, flow matching models not only produce
accurate solution fields for various PDE problems but also provide
uncertainty quantification through sampling-based predictions
\cite{li2503generative, chen2026flowmarchinggenerativepde,
kerrigan2023functional}. However, beyond unconditional or conditional
generation, other scientific tasks, such as out-of-distribution (OOD) detection, model
composition, and inverse problems, often require additional machinery at
inference time: flow matching allows exact likelihood computation but only through
expensive integration and trace estimation; guidance-based posterior sampling
does not faithfully target the posterior it defines \cite{Du2023-ms}; and for
inverse problems the required noisy likelihood must be approximated
\cite{Chung2022-qo}, for which the correction needs a different sampler \cite{wu2024practicalasymptoticallyexactconditional}.

Meanwhile, energy-based models (EBMs) \cite{LeCun2006-mt, Kim2016-ry,
Carbone2024-wu} as a family of generative models hold the promise of a single
model for all three tasks, due to their access to explicit energy functions.
However, energy-based models are challenging to train, often requiring expensive
sampling using MCMC and extensive hyperparameter tuning \cite{Du2019-kl,
Song2021-xv, Nijkamp2019-gu}. While methods such as denoising score matching
\cite{Vincent2011-zj} do not need expensive negative samples from the model,
they suffer from ill-defined scores and estimation difficulty outside of the
data support \cite{Song2019-vb}, and do not address the sampling obstacles where
noise-initialized MCMC chains have poor mixing, especially in high-dimensional
space \cite{koehler2022statisticalefficiencyscorematching}.
%\AS{Can we tune the above
%to focus on the limitations of EBMs and Flow Matching that we are alleviating?}

To combine the training and sampling efficiency of flow matching with the
explicit unnormalized density of EBMs, several hybrid formulations have recently
been proposed \cite{Balcerak2025-iy, Loo2025-nv, Zhou2025-cu}. Each requires a
customized training scheme rather than the standard flow matching objective, and
all remain under-explored for modeling PDE fields. In this work, we exploit the
explicit connection between flow matching models and EBMs by equating the score
functions of the marginal distributions at all transport times in flow matching
to the negative gradient of the energy function. Using a potential-induced
velocity parameterization within standard flow matching training, we recover
the time-dependent energy functions as a direct read-out. These energy functions,
as unnormalized negative log densities (Figure \ref{fig:demo}), grant a
single trained flow matching model the flexibility to perform efficient data
generation, OOD detection, and inverse problems. In particular, for modeling PDE
fields, we propose a total energy formulation that additively incorporates known
physical structures into the energy function to augment the learned prior. This
total energy admits an effective scoring function for OOD detection and also
improves sampling in data generation and inverse problems.
We summarize our contributions as follows:
\begin{itemize}
    \item We adopt a potential parameterization of standard flow matching that
    provides time-dependent EBMs along the transport path, making the converted
    score conservative by construction, enabling straightforward training and
    supporting multiple inference tasks.
    \item We construct a total energy by augmenting the data energy with known
    physical structures, and propose an Energy-based Predictor-Corrector
    (EnergyPC) framework that enables physics-aware unconditional generation and
    posterior sampling.
    \item We show that the data energy and the physical energy are complementary
    scoring functions for OOD detection, and that the total energy combining them
    is effective across the tested corruptions.
    \item We demonstrate the effectiveness of our approach on high-dimensional
    PDE datasets (Poisson, Helmholtz, and Burgers) through improved data
    generation, OOD detection, and PDE solution field reconstruction and
    coefficient inference compared to flow ODE, predictor--corrector, and
    guidance-based posterior sampling baselines.
    
    %\item Using high-dimensional PDE datasets (Poisson, Helmholtz, and Burgers), we show that EnergyPC improves physical data generation and posterior sampling for inverse problems compared to baseline flow matching and diffusion samplers.
    
    % allows energy-based correctors which can be incorporated in data
    % generation under the predictor-corrector framework to
    % augment the physical structure through prior energies to improve probabilistic
    % modeling of PDE fields.
    
    % \item based on a physics-augmented energy prior, improving physical data generation and posterior sampling for inverse problems in PDE fields.
    
    % \item We propose energy-based correctors which can be incorporated in data
    % generation and posterior sampling of PDE fields, effectively improving
    % sample quality and physical feasibility.
 
\end{itemize}

% Considering score-based diffusion models \cite{Song2020-iv} when using a
% conditional Gaussian path, velocity field in flow matching has an explicit relationship with the marginal scores in diffusion models, where we can view the marginal scores as negative gradient of the energy function In particular, we can model the
% time-marginals along the transport path using time dependent EBMs where the
% negative gradient of the energy function equals the marginal scores. In this
% work, we exploit this relationship and obtain noise-level dependent energy based
% models directly from a potential flow matching model and demonstrate the
% resulting energy functions are immediately applicable to be composed with other
% energies, e.g., physics constraints, to perform improved sampling, enable
% out-of-distribution detection, and solving inverse problems.
% \AS{Can we focus on how our approach alleviates the limitations of Flow matching and EBMs?} % not done with this par yet. will rewrite a lot

% \AS{List of contributions? } %-- will add those, too many meetings today very annoying..:(

\section{Related Work}
\paragraph{Energy-based models.}

Probabilistic energy-based models \cite{LeCun2006-mt, ngiam2011learning,
ackley1985learning} use the Boltzmann distribution to approximate the target
distribution through a parameterized energy function. EBMs have appealing
properties arising from the explicit access to the unnormalized density, making them a
natural choice for density estimation, out-of-distribution detection \cite{Liu2020-vz,
Wu2024-uo, Grathwohl2019-fa, Habring2025-lw}, model composition \cite{Du2024-ka,
Wu2023-ht}, and uncertainty quantification \cite{fuchsgruber2024energy,
friedli2025energy}. However, EBMs have historically been difficult to train and sample from,
requiring expensive Markov chain Monte Carlo sampling \cite{hinton2002training}
or its variants \cite{hinton2002training, Tieleman2008-jz, Du2019-kl}. Score
matching eliminates the need for expensive MCMC but faces challenges such as
ill-defined scores and estimation difficulties in areas off the low-dimensional
data manifold \cite{Song2019-vb}. Multilevel EBMs and noise annealing
substantially improve the learning of EBMs \cite{Song2019-vb, Gao2020-or, Li2019-qf}, but
they still require careful noise schedule design. Our work is closely related
to the multilevel treatment of EBMs, where we construct a series of EBMs at
different noise levels using a time-dependent energy function, but with a
straightforward noise schedule from the flow matching training objective.

\paragraph{Flow Matching.} % for Anirban and Sandeep
% \ys{mention that you get density using change of variables using the flow but point out the flaw: expensive, not suitbale for flexible composition, etc}
Flow matching \cite{Lipman2022-fd, Tong2023-yf} models attempt to learn a
velocity field, parameterized by a neural network, that transports samples from a
source (easy to sample) to a target (complex) data distribution. The learned
velocity is then used in either an ODE or SDE, initialized with random draws from
the source distribution and solved numerically to draw samples from the target
distribution. A well-trained flow matching model learns the ground truth marginal
distributions along the transport path \cite{liu2022flowstraightfastlearning}.
Despite their success in generative modeling, flow matching models do not
explicitly provide access to the likelihood or energy of the underlying data
distribution. The likelihood can be accessed using the change-of-variables
formula \cite{Lipman2022-fd, Chen2018-xz}, which is expensive for
high-dimensional datasets arising from physical processes. Flow matching has also
been applied to physical fields, with physics imposed through hard constraints,
guidance, or fine-tuning of the learned velocity \cite{Cheng2024-go, Utkarsh2025-gd, tauberschmidt2026physicsconstrainedfinetuningflowmatchingmodels, liang2025chanceconstrainedflowmatchinghighfidelity}. Our work instead uses
a potential parameterization, whose gradient approximates the velocity field, and
follows the standard flow matching training objective. This allows the
construction of marginal energy functions, which incorporate physical structure
additively and score OOD samples directly.
%\ys{add that we follow the flow matching training objecitve but obtian EBMs for OOD which is more efficient}

\paragraph{Energy-based flow/diffusion models.} Thanks to the common connection
to the score function, many works explore the relationship between flow matching
(or diffusion) models and EBMs. For example, Energy Matching
\cite{Balcerak2025-iy} uses a global time-independent potential flow to
transport samples from source distributions to near the target data manifold and
then runs MLE training via MCMC to obtain the global energy. VAPO
\cite{Loo2025-nv} creates a density-weighted Poisson equation and utilizes deep
Ritz method to solve the variational form of the equation as the training
objective to obtain the Boltzmann energy. Equilibrium matching
\cite{Wang2025-fk} learns an implicit energy based model by enforcing equilbrium
point at the end of the flow transport. While leveraging simplicity and strength
of flow matching, these frameworks need customized training target to attain the
EBM. Our work aligns closely with the energy diffusion models \cite{Du2023-ms,
Du2024-oe, Thornton2025-al} but under the flow matching framework for modeling
PDE fields, where we access the \emph{time-dependent} marginal energies through
a parameterized potential that induces the flow and augment the data-learned
prior energy with known physical structures.

\section{Methods}

The objective of our work is to learn the transport between the source Gaussian
$\mathcal{N}(0, I)$ and the target distribution of PDE fields $p_{\rm data}(x)$
through flow matching. Additionally, along the transport path, we are interested
in modeling the marginal distributions as EBMs and learn the energy functions.
We show in this section when parameterizing the flow velocity field using a
scalar potential, the energy function of marginal distributions can be directly
obtained in terms of the flow potential and the predefined noise schedule. Next,
we augment the flow matching-trained energies for PDE fields with known physical
structures, which enables energy-based correctors during sampling and admits a suitable 
scoring function for OOD detection.

\paragraph{Energy and Potential Flow.}

Let $x \in \mathbb{R}^d$, and let $p_0(x) = \mathcal{N}(0,I)$ and $p_1(x) =
p_{\rm data}(x)$ denote the source and target distributions. 
% both assumed to have
% well-defined densities with respect to the Lebesgue measure. 
Given the
interpolant $x_t = \alpha_t x_1 + \sigma_t x_0$, let $p_t(x)$ denote the induced
marginal at $t \in [0,1]$. We construct a series of \emph{time-dependent}
energy-based models such that
\begin{equation}\label{eq:ebm}
    p_t(x) = \frac{\exp(-E(x,t))}{Z_t}, \; Z_t = \int_{\mathbb{R}^d} \exp(-E(x,t))\,dx,
\end{equation}
where $E: \mathbb{R}^{d} \times [0,1] \to \mathbb{R}$, so that the score of the
time marginal is $\nabla_x \log p_t(x) = -\nabla_x E(x,t)$.

Consider a forward diffusion process transporting data samples to the source,
\begin{equation}\label{eq:forward_diff}
	dx_s = f(x,s)ds + g(s)dW_s, \; s: 0 \to 1,
\end{equation}
with $s = 1-t$. Throughout we use the linear path $\alpha_t = t$,
$\sigma_t = 1-t$, for which $f(x,1-t) = -x/t$ and $g^2(1-t) = 2(1-t)/t$ (see the
supplementary material). Reversing \eqref{eq:forward_diff} and substituting $t$
back \cite{anderson1982reverse, Song2020-iv}, the associated probability flow ODE is
\begin{equation}\label{eq:pf-ode}
    dx_t = \left[-f(x,1-t) + \tfrac{1}{2} g^2(1-t)\nabla_x \log p_t(x)\right]dt.
\end{equation}
Let the drift in \eqref{eq:pf-ode} be the gradient of a time-dependent potential
$\Phi(x,t)$, so that $dx_t/dt = \nabla_x\Phi(x,t)$ defines a flow
$\phi_t(x_0) = x_t$. Note that this formulation restricts the velocity to
curl-free fields; for a Gaussian path this is no restriction at the optimum,
since $f$ is linear and isotropic in $x$ and $g$ depends only on $t$, making both
terms of the \eqref{eq:pf-ode} drift gradients. The associated continuity equation
\begin{equation}\label{eq:continuity}
	\frac{\partial p_t}{\partial t} = -\nabla_x \cdot (p_t \nabla_x \Phi)
\end{equation}
shares the same marginal density as that induced by \eqref{eq:forward_diff}.
Solving \eqref{eq:pf-ode} for the score gives an explicit relationship between the
time marginals and the potential-induced velocity field,
\begin{equation}\label{eq:score-potential}
	\nabla_x \log p_t(x) = \frac{2\nabla_x \Phi(x,t) + 2f(x,1-t)}{g(1-t)^2},
\end{equation}
and combining with \eqref{eq:ebm} yields the energy
\begin{equation}\label{eq:marginal_energy}
	E(x,t) = \frac{\Vert x \Vert^2 - 2t\Phi(x,t)}{2(1-t)} + c(t),
\end{equation}
where $c(t)$ is a constant of integration with respect to $x$. We adopt
$c \equiv 0$, for which $E(x,0) = \Vert x \Vert^2/2$ recovers the energy of the
Gaussian source.

This energy function is the direct read-out from the potential that induces the
flow transporting Gaussian samples to the target distribution, and setting $c(t)=0$ does not affect fixed-time MCMC sampling and OOD scoring. Note that at
transport terminal time $t=1$, the energy becomes an indeterminate and numerically unstable representation.
% \SM{Eq. (10) is an indeterminate and numerically unstable representation at t=1; whether the true terminal energy exists depends on regularity and absolute continuity of the target distribution.}. 
In practice,
following the standard remedy \cite{Karras2022-iy,
ma2024sitexploringflowdiffusionbased} we set a minimum noise level
$\sigma_{\min}>0$ and treat the energy at that noise level as a close
approximation to the true data energy.

%\ys{comment on the regularity of the implicit potential to be learned}
% \AS{comment on the infinite energy when $t \to 1$? Look at Ma et al. (SiT paper) for comment on this issue.}
\paragraph{EBM training through Flow Matching.}
To obtain the energy function in (\ref{eq:marginal_energy}), we shift the learning objective from the velocity field to the potential function.
We parameterize the potential function $\Phi_\theta$ using a neural network with
the parameter set $\theta$, and we take the gradient, which is easily computable via automatic differentiation. Using the linear interpolant \cite{Lipman2022-fd, liu2022flowstraightfastlearning},
%\AS{call this I-CFM (tong et al.)?}
where $x_t =
(1-t)x_0 + tx_1, \; x_0 \sim \mathcal{N}(0, I), \ x_1 \sim p_{\rm data }(x)$, we obtain the standard flow
matching objective
\begin{equation}
    \mathcal{L}(\theta) = \mathbb{E}_{t, x_0, x_1}[\Vert \nabla_{x_t} \Phi_\theta(x_t,t) - (x_1 - x_0)\Vert_2^2].
\end{equation}
Once the model is trained, we obtain the flow potential $\Phi_\theta(x,t)$ along
with the learned energy functions $E_\theta(x,t)$ for downstream tasks. Note that
(\ref{eq:marginal_energy})
%\SM{refer to the correct equation}
is an identity at the population optimum, where the target marginal, the
marginal induced by $\nabla_x\Phi_\theta$, and the read-out density
$\exp\{-E_\theta\}/Z_\theta$ coincide. Under finite training they need not, as
(\ref{eq:marginal_energy}) assigns the energy pointwise without enforcing consistency with the
transport. We therefore treat $E_\theta$ as a read-out that inherits the accuracy
of the learned velocity, the same plug-in status as the velocity-to-score conversion in
flow-based SDE and posterior samplers \cite{Kim2025-sa, ma2024sitexploringflowdiffusionbased}. 
Also, with a standard neural network implementation (e.g., affine maps, Lipschitz
activations, and normalization layers), $\Phi_\theta$ grows at most linearly in
$\Vert x \Vert$, so the energy in (\ref{eq:marginal_energy}) yields a normalizable
density for $t<1$, admitting a valid EBM.
Such training of the EBMs takes advantage of the established matching objective, avoiding the expensive and unstable MCMC sampling required by conventional EBM training.

% \ys{comment on the velocity approixmation error and how that impact the energy readout}
% In turn, the flow matching formulation enables smooth transitions
% between energy landscapes at different noise levels, which leads to better
% calibrated data energy landscape.

\paragraph{Sampling.}
% The trained potential offers the marginal EBMs as well as their sampler. The
% potential flow transports samples from the source distribution to the time
% marginals by solving the flow ODE. Meanwhile, the explicit energy allows more
% flexible sampling strategies, such as MCMC and SMC. These approaches can either
% improve the ODE sampler by a predictor-corrector \cite{Song2020-iv} mechanism or
% completely bypass the flow dynamics and use annealing through the noise levels
% to achieve sampling.
%
% We show that the EBM from potential flow matching training is equipped with the
% flow ODE sampler and also offers general MCMC sampler in the predictor-corrector
% framework \cite{Song2020-iv}. The results show the $w_2$ and physics residual
% $R$ using the ODE sampler, total energy MALA refinement, MALA gated PC with
% total energy, and the annealed sampling on three pde datasets. Note
% that any MCMC techniques can be used here, we use MALA as an example.
Following the training outline above, a trained potential $\Phi_\theta$ admits
both the velocity field $\nabla_x\Phi_\theta(x,t)$, used for sampling by solving
the ODE, and the energy function $E_\theta(x,t)$, which allows MCMC-based
samplers to be incorporated. In addition, for general PDE fields, it is
straightforward to add known physical structures to the learned energy function
to refine the prior energy. We consider the PDE residual as the additional
structure and construct the total energy as
\begin{equation}\label{eqn:total-E}
E_{\rm tot}(x,t) = E_\theta(x,t) + \lambda_t \Vert R(x)\Vert^2,
\end{equation}
where $\Vert R(x)\Vert$ is squared the $L_2$ norm of the PDE residual,
corresponding to a Gaussian likelihood on the residual with $\lambda_t =
\frac{1}{2 \sigma^2_R(t)}$ and $\lambda_t$ is a time-dependent non-negative
scalar controlling its contribution. We present $E_{\rm tot}$ as an annealing
family of inference-time targets rather than the exact intermediate marginals
induced by the interpolant.  The residual of a noisy field is uninformative, so
we schedule $\lambda_t$ to zero at high noise levels. Therefore, the
intermediate members are not required to be the marginals of any distribution,
and serve only as an annealing path along which the corrector tracks the target.
At the terminal time the field is nearly clean, and the terminal member recovers
the learned prior corrected by the known physical structure.

Access to this total energy enables
more general MCMC samplers to be used in the predictor-corrector (PC) framework
introduced in \cite{Song2020-iv}, which is effective at limiting error
accumulation and improving distribution alignment when simulating the flow ODE.
Moreover, a corrector that uses $E_{\rm tot}$ additionally encourages the
physical feasibility of samples, which may not be captured by purely data-driven
training. We use $E_{\rm tot}$ in the corrector step and name the resulting
sampling process EnergyPC.

\newcommand{\Etot}{E_{\mathrm{tot}}}   % in preamble

\begin{algorithm}[t]
\caption{\textsc{EnergyPC}: energy-based predictor--corrector with optional adjustment}
\label{alg:epc}
\begin{algorithmic}[1]
\Require transport-time grid $0 = t_0 < \cdots < t_N = 1-\sigma_{\min}$, corrector steps $M$,
         noise floor $\sigma_{\min}$, ODE solver \textsc{ODEStep},
         Langevin proposal $\mathcal{Q}$ with step size $\eta$, flag \textsc{adjust}
\Ensure approximate sample from $p_1$
\State $x \sim \mathcal{N}(0, I)$ %\Comment{source distribution at $t=0$}
\For{$k = 0, \dots, N-1$}
  \State $x \gets \textsc{ODEStep}\big(x,\ \nabla_x \Phi_\theta,\ t_k \to t_{k+1}\big)$
         % \Comment{predictor: one step of $\dot{x} = v_\theta(x,t)$}
  \State $\sigma \gets 1 - t_{k+1}$
  \For{$j = 1, \dots, M$}
    \State $(\hat{x}, \Lambda) \gets \mathcal{Q}.\textsc{Propose}\big(x,\ \nabla_x E_{\rm tot}(x, t_{k+1}),\ \eta\big)$,
           where $\Lambda = \log q(x \,|\, \hat{x}) - \log q(\hat{x} \,|\, x)$
    \If{\textsc{adjust}}
      \State $\log\alpha \gets \Etot(x,t_{k+1}) - \Etot(\hat{x},t_{k+1}) + \Lambda$
      \State $u \sim \mathcal{U}(0,1)$;\quad
             $x \gets \hat{x}$ \textbf{if} $\log u < \log\alpha$ \textbf{else} $x$
    \Else
      \State $x \gets \hat{x}$ %\Comment{unadjusted: always accept}
    \EndIf
  \EndFor
\EndFor
\State \Return $\hat{x}_1 \gets x + \sigma_{\min}\, \nabla_x \Phi_\theta(x, 1-\sigma_{\min})$
       %\Comment{Tweedie estimate $\mathbb{E}[x_1 \mid x_{t_N}]$}
\end{algorithmic}
\end{algorithm}

Algorithm \ref{alg:epc} shows the EnergyPC framework that uses the Metropolis
adjusted Langevin algorithm (MALA) as the MCMC corrector. Note that the
energy-based PC framework also accepts other samplers that utilize energy, such
as Hamiltonian Monte Carlo, replica exchange, and Jarzynski-reweighted variants \cite{earl2005parallel, jarzynski1997nonequilibrium}.

% \ys{should
% specify that the standard PC uses pure scores alone so energyPC with ULA is
% considered standard PC}

At each predictor step $k$ and transport time $t$, we run $M$ 
%\SM{Algorithm 1 uses $M$ corrector steps} 
steps of corrector, for which the proposal
is given by a Langevin step on the total energy,
\begin{equation}
\begin{aligned}
    \hat{x}_k = x_k - \eta \nabla_x E_{\rm tot} + \sqrt{2\eta} \epsilon,\; \epsilon \sim \mathcal{N}(0, I)\\
\end{aligned}
\end{equation}
Then a Metropolis adjustment step decides whether to accept the proposal,
\begin{equation}
    \alpha = \min \left\{1, \frac{p_t(\hat{x}_k)q(x_k \mid \hat{x}_k)}{p_t(x_k)q(\hat{x}_k \mid x_k)}\right \},
    % q(x'\mid x) = \mathcal N!\big(x';, x+\delta(x)s(x),; 2\delta(x)I\big)
\end{equation}
where $q(\hat{x}_k \mid x_k) = \mathcal{N}(\hat{x}_k; x_k -\eta \nabla_x E_{\rm tot}(x_k, t), 2\eta I)$ and $q(x_k \mid \hat{x}_k) = \mathcal{N}(x_k; \hat{x}_k -\eta \nabla_{\hat{x}} E_{\rm tot}(\hat{x}_k, t), 2\eta I)$. Acceptance requires
$\alpha \geq u, \; u \sim U[0, 1] $. Here, $\alpha$ is easily computable through the energy function as $p_t(x) \propto \exp(-E_{\rm tot}(x, t))$. 
Compared to the unadjusted Langevin step used in the standard PC framework, the
Metropolis adjustment makes the corrector kernel exactly invariant to
$\exp\{-E_{\rm tot}(x,t)\}$ at each noise level, removing the discretization bias
of the unadjusted step \cite{roberts1996exponential}.

% \input{manuscript/energy_pc_algo}
% here goes the algos

\paragraph{Energy as Scoring Function.} Energy functions measure the
compatibility of the data with underlying distribution, where the compatible
data with higher likelihood present low energy values. Therefore, energy
functions provide suitable scoring for OOD tasks. Similar to the settings in \cite{Liu2020-vz, Wu2024-uo},
we construct a classifier from the terminal time energy $E_{\rm tot}(x, t_{\max}), \ t_{\max}= 1 - \sigma_{\min}$ to perform inference
time OOD using a discriminator defined as
\begin{equation}
    D(x) = \begin{cases}
        0, \ E_{\rm tot}(x, t_{\max}) \leq \tau\\
        1, \ E_{\rm tot}(x, t_{\max}) > \tau
    \end{cases},
\end{equation}
where $\tau$ is the scoring threshold, and an input of energy value greater than
the threshold is considered an out-of-distribution sample, as a lower energy
corresponds to a higher likelihood in (\ref{eq:ebm}). 

%\AS{appendix?}
We use AUROC \cite{hendrycks2016baseline} to evaluate the performance of the
energy function as a scoring function. Using $N_{\rm in}$ in-distribution test
fields and an equal number $M=N_{\rm in}$ of corrupted fields per tier, we
compute the energy values of these samples and denote them as $s_i, \ i=1,\dots,
N_{\rm in}$ and $t_j, \ j= 1, \dots, M$. AUROC, defined in (\ref{eqn:auroc}), measures the fraction of
in-distribution and out-of-distribution pairs the energy ranks correctly,
therefore represents the probability of the energy score correctly ranking such
pairs.
\begin{equation}\label{eqn:auroc}
  \text{AUROC} = \frac{1}{N_{\rm in}M} \sum_i \sum_j [ \mathds{1}(t_j > s_i) +
\frac{1}{2}\mathds{1}(t_j = s_i) ]
\end{equation}
Here, a value of 1 indicates that the OOD samples are perfectly separable using
the energy score, while a value smaller than 0.5 means the OOD samples appear
more in-distribution than real samples, which implies score failure that the
energy carries the opposite information. A value of 0.5 means the energy carries
no information to rank the sample pairs.

\paragraph{Posterior Energy for Inverse Problem.}
The learned energy from flow matching training is useful in solving inverse
problems, since EBMs provide a straightforward definition of the posterior by
adding the likelihood energy term. For PDE fields with an $(a,u)$ pair, where $a$
is the coefficient and $u$ is the solution field, let $\{o_i\}_{i=1}^N$ denote a
set of observed points in the solution field, related to the field through the
observation operator $H$ as
\begin{equation}\label{eqn:likelihood}
    o = H(x) + \epsilon, \ \epsilon \sim \mathcal{N}(0, \sigma^2_{\rm obs} I).
\end{equation}
The likelihood is then $p(o \mid x) = \mathcal{N}(o; H(x), \sigma^2_{\rm obs}I)$,
and the posterior of the joint field $x = (a,u)$ is
\begin{equation}
\begin{aligned}
    \log p(x|o) &= \log p(x) + \log p(o\mid x) + C\\
    & = -\underbrace{\left(E_{\rm tot} + \frac{\Vert \{o_i\}_{i=1}^N - H(x)\Vert^2}{2\sigma^2_{\rm obs}}\right)}_{\text{posterior energy} \, = \, E_{\rm pos}} + C,
\end{aligned}
\end{equation}
where $C$ collects the normalizing constants of the prior and likelihood terms.
The likelihood energy from the conditional Gaussian distribution
(\ref{eqn:likelihood}) is quadratic and composes directly with the prior energy.
As with $E_{\rm tot}$, $E_{\rm pos}$ defines an annealing family rather than the
sequence of marginals the flow dynamics would transport. Here, the quadratic likelihood
is evaluated on the current field at every noise level, which is accurate only as
the field becomes clean, and the terminal member coincides with the flow posterior.

Similar to the use of energy in the PC framework, existing flow-based posterior samplers (e.g.,
FlowDPS \cite{Kim2025-sa}) can leverage the posterior energy to improve accuracy.
At each posterior sampling step, the guided velocity with guidance coefficient $\nu$ at noise level $t$ is
\begin{equation}
 \nabla_x \Phi_\theta(x, t \mid o) = \nabla_x \Phi_\theta(x, t) + \nu \nabla_x \log p(o \mid \mathbb{E}[x_1 \vert x_t]),
\end{equation}
and we apply the energy-based MCMC corrector (step 5-11 in Algorithm \ref{alg:epc}) at the same noise level
\begin{equation}\label{eqn:energy-corr}
    x \leftarrow \texttt{EnergyCorr}(x, E_{\rm pos}, \eta).
\end{equation}
The posterior energy gives the corrector an explicit target at each noise level,
which the guidance term alone does not provide.

% should i comment on that the energy definedd posteriro does not need tweedie estimation of the clean but create an annealing target/

\section{Experiments}
We train and evaluate the potential flow matching framework on the PDE field
datasets from \cite{Huang2024-rv} and show the impact of the energy functions in
unconditional generation, OOD detection, and inverse problems on solution field
reconstruction and coefficient inference. Model architecture, hyperparameters, and
training details are described in the supplementary material.

% \ys{Potential flow training. energy formulation (dot product) and the training steps; defer details to the appendix.}

\paragraph{Unconditional generation}

% here is the summary table
\begin{table*}[t]
\centering
\small
\caption{Unconditional generation metrics across samplers and datasets, with 95\% bootstrap
confidence intervals (2000 resamples of the 100-sample ensemble).
MMSE/SMSE ${\times}10^{-4}$; Burgers PDE residual ${\times}10^{-4}$. \textbf{Bold} marks the
best mean value per column within each dataset.}
\label{tab:uncond_all_ci}
\begin{tabular}{llcccccc}
\toprule
Dataset & Sampler & NFE & Sliced-$W_2\!\downarrow$ & Spectral$\downarrow$ & PDE resid$\downarrow$ & MMSE$\downarrow$ & SMSE$\downarrow$ \\
\midrule
\multirow{6}{*}{\shortstack[l]{Poisson}}
 & ODE120                 & 120 & $0.0532{\pm}0.0046$ & $1.180{\pm}0.079$ & $13.66{\pm}5.46$ & $2.66{\pm}3.14$ & $16.33{\pm}4.30$ \\
 & ODE240                 & 240 & $0.0520{\pm}0.0048$ & $1.181{\pm}0.079$ & $14.59{\pm}5.91$ & $2.92{\pm}3.51$ & $12.72{\pm}3.75$ \\
 & ODE360                 & 360 & $0.0518{\pm}0.0048$ & $1.181{\pm}0.076$ & $14.94{\pm}5.82$ & $3.01{\pm}3.71$ & $\mathbf{11.58{\pm}3.69}$ \\
 & Standard PC            & 240 & $0.0449{\pm}0.0047$ & $1.013{\pm}0.034$ & $4.54{\pm}0.93$ & $3.12{\pm}2.73$ & $22.51{\pm}4.91$ \\
 & EnergyPC-ULA (ours)    & 240 & $0.0449{\pm}0.0050$ & $0.979{\pm}0.024$ & $3.65{\pm}0.57$ & $3.12{\pm}2.85$ & $22.51{\pm}4.87$ \\
 & EnergyPC-MALA (ours)   & 360 & $\mathbf{0.0420{\pm}0.0050}$ & $\mathbf{0.915{\pm}0.032}$ & $\mathbf{2.46{\pm}0.64}$ & $\mathbf{1.81{\pm}2.06}$ & $18.35{\pm}4.27$ \\
\midrule
\multirow{6}{*}{\shortstack[l]{Helmholtz}}
 & ODE120                 & 120 & $0.0594{\pm}0.0038$ & $0.919{\pm}0.055$ & $2.86{\pm}1.49$ & $\mathbf{4.12{\pm}3.03}$ & $28.40{\pm}4.77$ \\
 & ODE240                 & 240 & $0.0561{\pm}0.0038$ & $0.911{\pm}0.061$ & $3.13{\pm}1.65$ & $4.50{\pm}3.52$ & $23.07{\pm}4.47$ \\
 & ODE360                 & 360 & $\mathbf{0.0550{\pm}0.0039}$ & $0.909{\pm}0.059$ & $3.24{\pm}1.59$ & $4.65{\pm}3.79$ & $\mathbf{21.28{\pm}4.46}$ \\
 & Standard PC            & 240 & $0.0574{\pm}0.0040$ & $0.942{\pm}0.022$ & $3.69{\pm}0.51$ & $6.67{\pm}4.05$ & $30.53{\pm}5.14$ \\
 & EnergyPC-ULA (ours)    & 240 & $0.0574{\pm}0.0042$ & $0.942{\pm}0.018$ & $3.69{\pm}0.41$ & $6.69{\pm}4.25$ & $30.52{\pm}5.20$ \\
 & EnergyPC-MALA (ours)   & 360 & $0.0551{\pm}0.0038$ & $\mathbf{0.848{\pm}0.026}$ & $\mathbf{1.19{\pm}0.36}$ & $5.48{\pm}3.45$ & $27.74{\pm}5.00$ \\
\midrule
\multirow{6}{*}{\shortstack[l]{Burgers}}
 & ODE120                 & 120 & $0.0521{\pm}0.0147$ & $1.734{\pm}0.098$ & $3.59{\pm}0.69$ & $9.16{\pm}13.57$ & $12.07{\pm}9.96$ \\
 & ODE240                 & 240 & $0.0616{\pm}0.0155$ & $1.814{\pm}0.104$ & $4.21{\pm}0.82$ & $9.95{\pm}14.09$ & $21.82{\pm}14.29$ \\
 & ODE360                 & 360 & $0.0654{\pm}0.0158$ & $1.842{\pm}0.102$ & $4.46{\pm}0.85$ & $10.25{\pm}14.35$ & $26.12{\pm}15.17$ \\
 & Standard PC            & 240 & $0.0451{\pm}0.0118$ & $1.399{\pm}0.084$ & $3.36{\pm}0.62$ & $7.19{\pm}9.74$ & $11.43{\pm}8.13$ \\
 & EnergyPC-ULA (ours)    & 240 & $0.0451{\pm}0.0120$ & $1.399{\pm}0.089$ & $3.36{\pm}0.62$ & $7.19{\pm}9.81$ & $11.43{\pm}8.12$ \\
 & EnergyPC-MALA (ours)   & 360 & $\mathbf{0.0435{\pm}0.0123}$ & $\mathbf{1.303{\pm}0.089}$ & $\mathbf{3.05{\pm}0.52}$ & $\mathbf{4.64{\pm}8.02}$ & $\mathbf{8.50{\pm}6.91}$ \\
\bottomrule
\end{tabular}
\end{table*}

% additional -- poisson
% PC-ULA  (w_phys=0)  240  0.0449  1.013  4.54  3.12  22.51
% PC-MALA (w_phys=0)  360  0.0424  0.916  2.47  1.80  18.63

% helm
% ┌──────────────────────────┬────────┬────────┬───────┬─────────┬─────────┐
% │ Helmholtz PC-ULA (gated) │   W2   │  Spec  │ Resid │  MMSE   │  SMSE   │
% ├──────────────────────────┼────────┼────────┼───────┼─────────┼─────────┤
% │ w=1                      │ 0.0574 │ 0.9415 │ 3.692 │ 6.69e-4 │ 3.05e-3 │
% ├──────────────────────────┼────────┼────────┼───────┼─────────┼─────────┤
% │ w=0                      │ 0.0574 │ 0.9420 │ 3.693 │ 6.67e-4 │ 3.05e-3 │
% └──────────────────────────┴────────┴────────┴───────┴─────────┴─────────┘

%\AS{Can we specifically say which baselines we are comparing against?}
We compare the performance of unconditional generation of the PDE fields using
the proposed EnergyPC framework against the flow ODE sampler \cite{Lipman2022-fd,
Tong2023-yf} and the standard PC framework \cite{Song2020-iv} on the Poisson,
Helmholtz, and Burgers equation datasets. Note that EnergyPC-ULA can also be
considered the standard PC framework, as it requires only the gradient of the
energy function. We use $E_{\rm tot}(x, t)$ in the energy corrector step listed in
Algorithm \ref{alg:epc} and generate 100 samples from each sampling process.
Sample quality is assessed using sliced $W_2$ distance, radial log-spectral
distance, mean squared error of the mean (MMSE), mean squared error of the
standard deviation (SMSE), and PDE residuals (see the supplementary material for
details).
%\AS{[formulation in the appendix]}

\begin{figure}[t]
	\begin{center}
		\includegraphics[width=\linewidth]{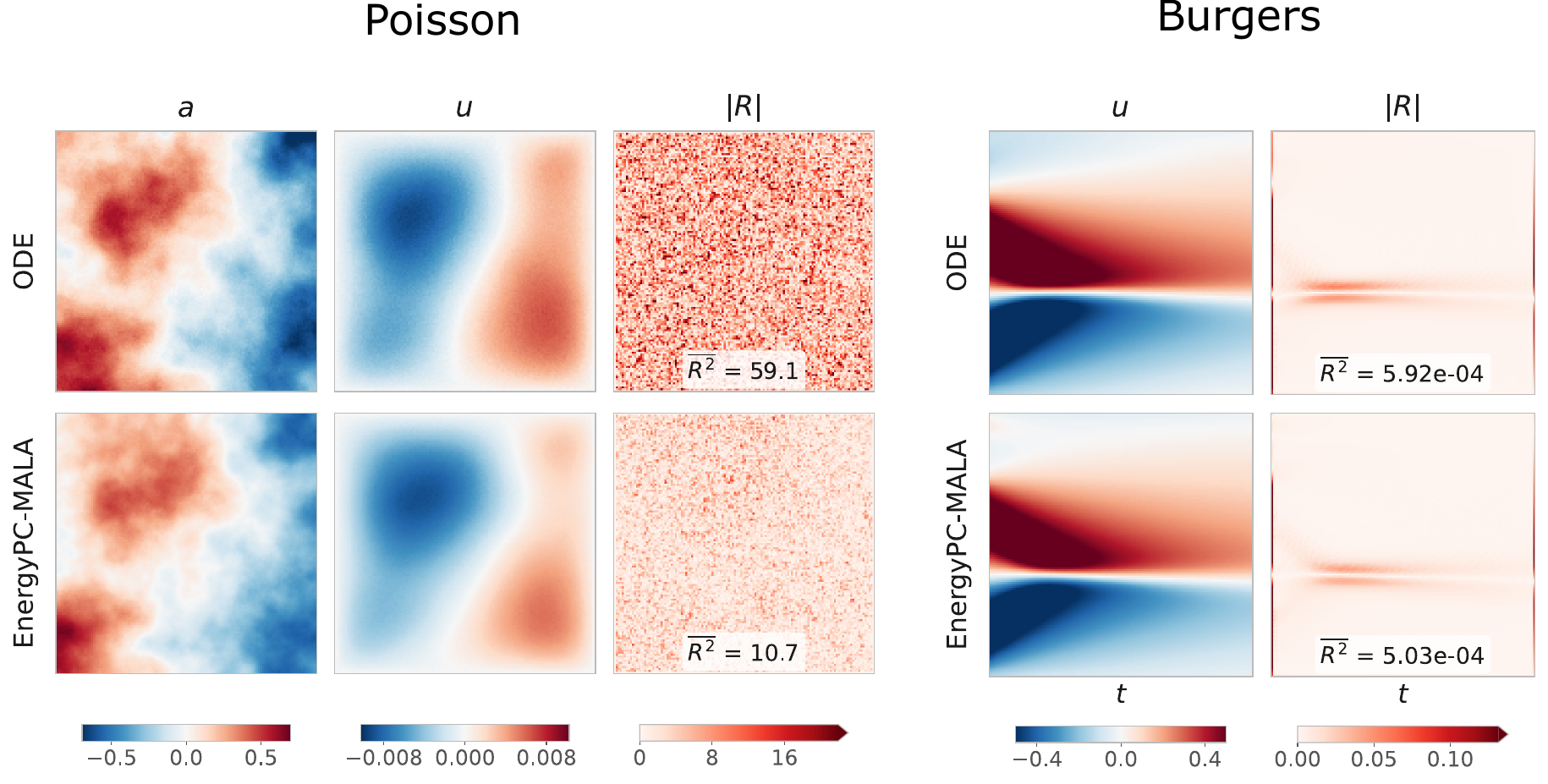}
	\end{center}
	\caption{Visualization of samples of Poisson and Burgers samples generated using the baseline flow ODE and EnergyPC with MALA. 
    %\SM{update the caption with Burgers equation} 
    }\label{fig:energypc}
\vspace{-1em}
\end{figure}

We conduct two sets of comparisons against the ODE sampler: one with the same
number of noise levels, where the EnergyPC framework consumes extra computation
per step, and the other with the same computational budget, where it uses a
coarser transport time grid. In particular, the ODE sampler runs at three
budgets: 120, 240, and 360 transport steps. Standard PC and EnergyPC samplers
use the same 120 predictor steps, with additional corrector cost matching 240
and 360 neural function evaluations for the ULA and MALA correctors,
respectively.

Table \ref{tab:uncond_all_ci} shows the generation metrics with different samplers
for the three datasets. We observe that the corrector with energy-based adjustment
(MALA) is superior to the purely score-based corrector (ULA), and that EnergyPC
with ULA outperforms the standard PC on Poisson and has similar performance on
Helmholtz and Burgers generation. Particularly, the application of EnergyPC with
MALA effectively reduces the PDE residual and spectral distance on \emph{all}
three datasets. Figure \ref{fig:energypc} shows the effect of EnergyPC visually,
where the refined samples present improved physical feasibility. Nevertheless,
unlike the physical feasibility, the EnergyPC framework has a less consistent
effect on the MMSE and SMSE metrics. This could be due to the drift-dominant
corrector step from the SNR-adjusted step size \cite{Song2019-vb} and the limited
sample size.

\begin{figure}[t]
    \centering
    \includegraphics[width=.9\linewidth]{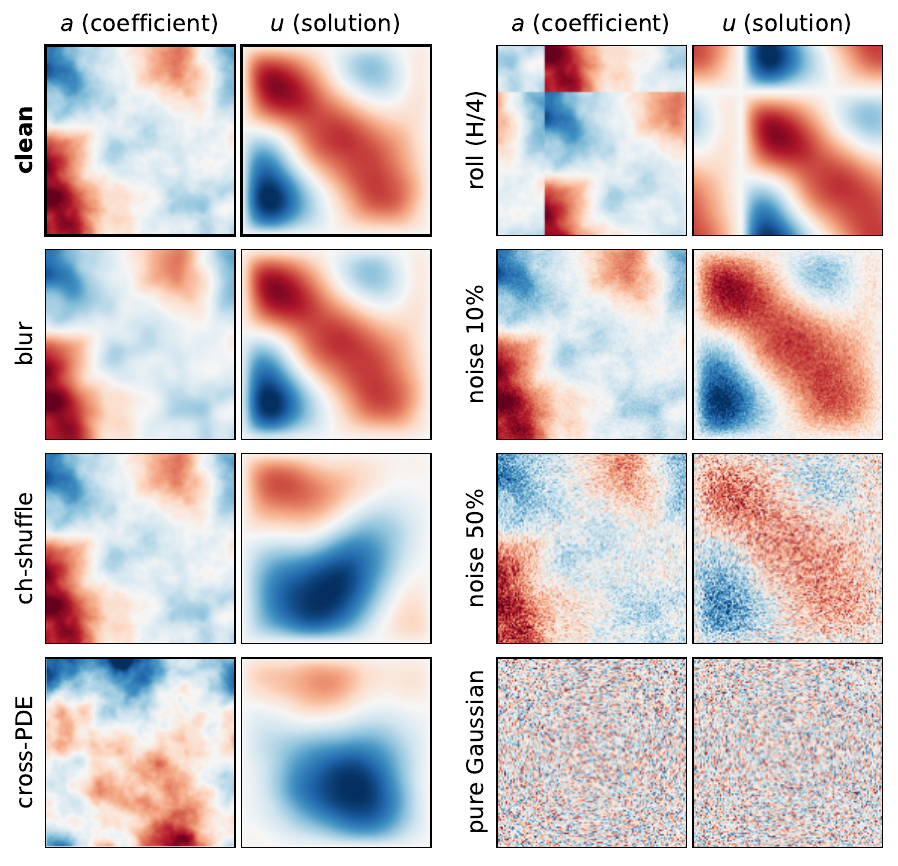}
    \caption{Synthetic OOD sample examples on the Poisson $a$ and $u$ pairs.}\label{fig:ood}
\vspace{-1em}
\end{figure}

% OOD detection: energy vs residual complementarity. Auto-generated.
\begin{table*}[t]\centering\small
\caption{OOD detection AUROC with the energy, with 95\% bootstrap confidence intervals
(\emph{italic}=at or near chance). The
learned energy $E$ and the PDE residual $R$ are complementary: each is blind where the other
separates, and the total energy $E_{\rm tot}$ flags every tier where either does.}
\label{tab:ood_ci}
\begin{tabular}{l ccc ccc ccc}\toprule
& \multicolumn{3}{c}{Poisson} & \multicolumn{3}{c}{Helmholtz} & \multicolumn{3}{c}{Burgers} \\
\cmidrule(lr){2-4}\cmidrule(lr){5-7}\cmidrule(lr){8-10}
OOD tier & $E$ & $R$ & $E_{\rm tot}$ & $E$ & $R$ & $E_{\rm tot}$ & $E$ & $R$ & $E_{\rm tot}$ \\\midrule
Gaussian noise            & 1.00 & 1.00 & 1.00 & 1.00 & 1.00 & 1.00 & 1.00 & 1.00 & 1.00 \\
Noise ($10\%$ field std)  & 1.00 & 1.00 & 1.00 & 1.00 & 1.00 & 1.00 & 1.00 & $0.843{\pm}0.027$ & 1.00 \\
Noise ($50\%$ field std)  & 1.00 & 1.00 & 1.00 & 1.00 & 1.00 & 1.00 & 1.00 & 1.00 & 1.00 \\
Channel shuffle           & $0.992{\pm}0.005$ & 1.00 & 1.00 & 1.00 & 1.00 & 1.00 & --   & --   & --   \\
Cross-PDE                 & 1.00 & 1.00 & 1.00 & 1.00 & 1.00 & 1.00 & --   & --   & --   \\
Blur / smoothing          & $\mathit{0.636{\pm}0.034}$ & 1.00 & 1.00 & 1.00 & 1.00 & 1.00 & $0.614{\pm}0.035$ & $\mathit{0.490{\pm}0.034}$ & $\mathit{0.672{\pm}0.032}$ \\
Spatial roll              & 1.00 & 1.00 & 1.00 & 1.00 & 1.00 & 1.00 & 1.00 & $\mathit{0.558{\pm}0.035}$ & 1.00 \\
\bottomrule\end{tabular}\end{table*}
~

\paragraph{Out-of-distribution detection}
We investigate the usage of $E_{\rm tot}$ in OOD tasks by comparing the energy
values of corrupted samples against those of true data samples. We synthesize OOD
samples from true data samples by applying blur, roll, channel shuffle, and noise
injection operations, and use samples across PDEs (Poisson and Helmholtz) as
additional OOD instances. The blur operation is a Gaussian convolution, which acts
as a low-pass filter on the original clean fields. The roll operation shifts the
fields, preserving the bulk physical feasibility. Channel shuffle
creates a mismatch between the coefficient and solution fields, so that $a$ and
$u$ are individually in-distribution but jointly out-of-distribution. Noising
operations add Gaussian noise scaled by the field standard deviation. Figure
\ref{fig:ood} shows example results of the above operations.

We report the AUROC values computed from $N_{\rm in} = M = 256$ true and OOD
samples, using energy terms in (\ref{eqn:total-E}) separately,
where $E_\theta(x,t)$ encodes the distributional information learned from data,
$\Vert R\Vert$ detects physical inconsistency, and $E_{\rm tot}$ combines the two
complementary terms, which the experiments show to be a better scoring function
for OOD detection.
Values in Table \ref{tab:ood_ci} indicate that in most of the testing cases, the
data energy and PDE residual alone are capable of perfect or near-perfect
detection of OOD samples, which means the separate energy terms are adequate
scoring functions that correctly rank true and OOD samples. In particular, all
energies capture the channel-shuffled cases, where each channel ($a$ and $u$)
individually satisfies the distribution. This demonstrates that the learned
energies score the channels jointly, where the mismatched pairs are off-manifold.

However, the Gaussian convolution blurred samples cause a large accuracy
reduction on the data energy for both the Poisson and Burgers datasets. The
smoothed samples are visually consistent with true samples despite the loss of
high frequency details, and lie towards the mode of the data distribution,
corresponding to high likelihood regions, but not in the distribution's typical
set \cite{Nalisnick2019-gt}. Hence, the data energy term alone struggles to
distinguish the blurred samples from true samples, while $E_{\rm tot}$ improves
the detection accuracy for both datasets. Similarly, due to the periodic boundary
conditions in the Burgers data, the spatial roll operation preserves the physical
feasibility of a sample but pulls it out of the data distribution. Thus, the
physical energy term alone could not distinguish the corrupted samples cleanly,
but $E_{\rm tot}$ prevents the failure.

% On the
% other hand, for the Poisson case, smoothed fields violate the physical
% structure, leading to an increase in the energy associated with PDE residuals.
% As a result, both $\Vert R\Vert$ and $E_{\rm tot}$ successfully detected the OOD cases. On the contrary, the Gaussian convolution blurring does not push
% Burgers PDE residual to a higher value, because the blur operation coincides
% with the equation's diffusion term \cite{evans2022partial}. \ys{but blur on
% burgers goes both x and t axes; needs a closer look.} Meanwhile, Burgers' PDE
% residual alone also suffers from the spatial roll corrupted samples as the
% Burgers dataset uses a periodic boundary condition and this operation does not
% break its physical structure. But such failure from using the PDE residual is
% addressed by the data energy term.

The experiment showcases that the learned data energy and PDE residual energy
are complementary as OOD scoring functions, and that the total $E_{\rm tot}$ is
a suitable single scoring function for OOD detection purposes.

\paragraph{Inverse Modeling.}
\begin{table}[t]\centering\small
\caption{Relative $L_2$ errors of reconstructed $u$ and inferred $a$ in the inverse problem.
Fields are the ensemble mean over 16 samples; energy correctors are added at the base's own
predictor levels. $\pm$ is the standard deviation over resamples of the 16 draws.}
\label{tab:inverse_samelevels}
\resizebox{\linewidth}{!}{
\begin{tabular}{llcc}\toprule
Dataset & Method & aRE $\downarrow$ & uRE$\downarrow$ \\
\midrule
\multirow{6}{*}{Poisson} & DiffusionPDE  & $0.285{\pm}0.020$ & $\mathbf{0.146{\pm}0.017}$ \\
 & \quad +$E_{\rm pos}$-ULA   & $0.304{\pm}0.057$ & $0.151{\pm}0.018$ \\
 & \quad +$E_{\rm pos}$-MALA  & $\mathbf{0.255{\pm}0.010}$ & $0.150{\pm}0.020$ \\
 & FlowDPS  & $0.378{\pm}0.008$ & $0.039{\pm}0.001$ \\
 & \quad +$E_{\rm pos}$-ULA   & $0.373{\pm}0.007$ & $0.038{\pm}0.001$ \\
 & \quad +$E_{\rm pos}$-MALA  & $\mathbf{0.353{\pm}0.012}$ & $\mathbf{0.036{\pm}0.001}$ \\
\midrule
\multirow{6}{*}{Helmholtz} & DiffusionPDE  & $0.254{\pm}0.008$ & $0.139{\pm}0.028$ \\
 & \quad +$E_{\rm pos}$-ULA   & $\mathbf{0.231{\pm}0.007}$ & $0.129{\pm}0.018$ \\
 & \quad +$E_{\rm pos}$-MALA  & $0.244{\pm}0.008$ & $\mathbf{0.128{\pm}0.030}$ \\
 & FlowDPS  & $0.338{\pm}0.009$ & $0.047{\pm}0.001$ \\
 & \quad +$E_{\rm pos}$-ULA   & $\mathbf{0.315{\pm}0.007}$ & $\mathbf{0.045{\pm}0.001}$ \\
 & \quad +$E_{\rm pos}$-MALA  & $0.319{\pm}0.007$ & $0.046{\pm}0.001$ \\
\bottomrule\end{tabular}}
\vspace{-1em}
\end{table}
We also test the role of energy functions in posterior sampling on top of popular
guidance-based diffusion posterior samplers. Following the inverse problem setup
in \cite{Huang2024-rv}, for the Poisson and Helmholtz datasets, we randomly sample
500 locations in the solution field $u$ as the observed points, and reconstruct
the solution field as well as infer the corresponding coefficient field $a$. We
evaluate model performance by incorporating the energy correctors into the
DiffusionPDE \cite{Huang2024-rv} and FlowDPS \cite{Kim2025-sa} frameworks.

\begin{figure}
    \centering
    \includegraphics[width=\linewidth]{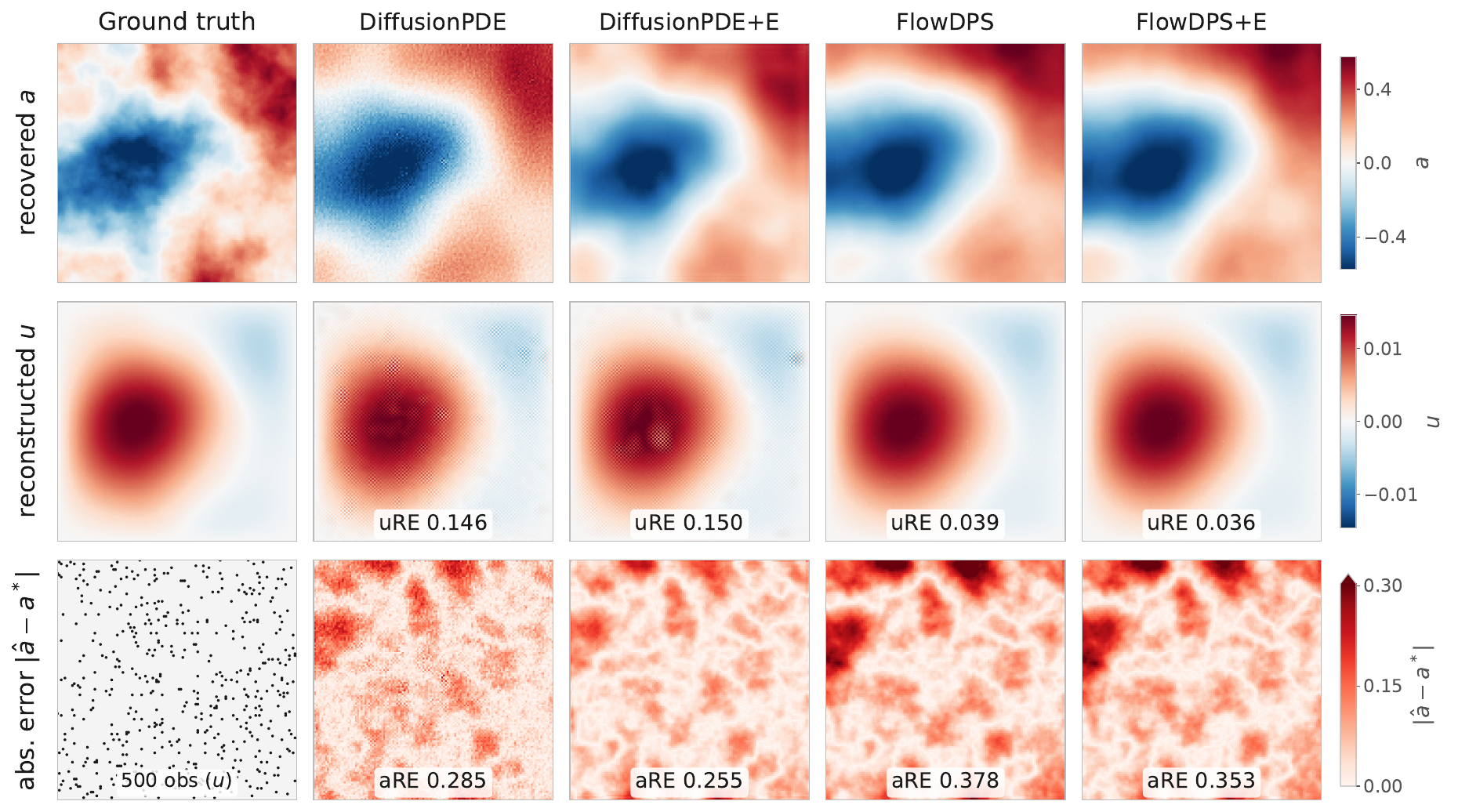}
    \caption{Poisson solution field reconstruction and coefficient inference comparison using energy-corrector with MALA (+E) extended diffusion posterior samplers. 
    %\SM{say what FlowDPS+E is}
    }
    \label{fig:inverse}
\vspace{-1em}
\end{figure}

Similar to the procedure outlined in Algorithm \ref{alg:epc}, we apply an
energy-based corrector step (\ref{eqn:energy-corr}) with the \emph{posterior
energy} after each guidance-updated ODE integration step. Table
\ref{tab:inverse_samelevels} shows the relative $L_2$ errors of the reconstructed
solution and inferred coefficient fields produced by DiffusionPDE, FlowDPS, and
their energy-corrector extensions, and Figure \ref{fig:inverse} shows an example
of the effect. Overall, the energy correctors lead to improved results within both
frameworks, with the largest and most consistent gains in coefficient inference;
the exception is DiffusionPDE on Poisson, where the solution field error increases
slightly. More specifically, for the Poisson equation data, MALA energy correctors
outperform ULA energy correctors in both the DiffusionPDE and FlowDPS settings. For
the Helmholtz data, while both correctors outperform the baseline, the ULA
corrector shows marginal improvement over the MALA corrector. Nevertheless, rooted
in the explicitly defined posterior energy, the energy correctors exhibit enhanced
performance in posterior sampling.

\section{Conclusion} We show in this work that a potential-parameterized flow
matching model with a Gaussian probability path provides explicit energy
functions for the transport-time marginal distributions. When modeling PDE
fields, such energies combined with additional physical structure create an
augmented prior, the total energy, which enables a flexible energy
adjustment-based predictor--corrector framework (EnergyPC) for refined sampling
that improves sample quality and physical feasibility. Concurrently, the total
energy, consisting of complementary terms for OOD scoring, can also act as an
effective single scoring function for OOD detection. Furthermore, the total
energy composes straightforwardly with the observational likelihood energy, which
leads to better field reconstruction and coefficient inference in PDE field
inverse problems.

\paragraph{Limitations.} In spite of the success in data
generation, OOD detection, and inverse problems, the proposed framework has two major limitations. First, the potential parameterization of the flow leads
to additional computational complexity compared to the standard flow matching
models, which will affect both training and sampling efficiency.
% \AS{Do we
% mention some of the challenges and how we alleviated them in this work in the
% main text or in appendix?}. 
Second, the read-out energy becomes singular algebraically at the
terminal transport time $t=1$. So, the true terminal energy is not recoverable directly from this framework. Future work will focus on addressing these challenges.
% Last, the EBM training through the flow matching objective
% restricts the energy landscape only shaped around noised data samples at
% different noise levels, which might cause challenges for pure MCMC-based
% samplers. \ys{add future work addressing these limitations}

\bibliography{aaai2027, paperpile, ref}

% Check whether the conference requires a reproducibility checklist to be included in the paper.
% If so, you can uncomment the following line and ajust the path to include it.

\newpage

\newcommand{\xhat}{\hat{x}_1}
\newcommand{\Rres}{R}

% \title{Supplemental Materials for Submission}
% \date{}
% \begin{document}
\onecolumn
\pagestyle{plain}
\appendix
\setcounter{secnumdepth}{2}

%======================================================================
\section{Energy formulation}
%======================================================================

\subsection{The identification of $f$ and $g$}
From the main text, the linear interpolant is
\begin{equation}
  x_t = t\,x_1 + (1-t)\,x_0, \qquad x_0 \sim \mathcal N(0, I), \qquad t: 0 \to 1,
  \label{eq:interp}
\end{equation}
so the conditional path is Gaussian, $p_t(x \mid x_1) = \mathcal N\!\big(\alpha_t x_1,
\sigma_t^2 I\big)$ with
\begin{equation}
  \alpha_t = t, \qquad \sigma_t = 1-t .
\end{equation}
Introduce the diffusion time $s = 1-t$, so $s: 1 \to 0$ is the forward
direction corresponding the transport from data to source, and $\alpha(s) = 1-s$, $\sigma(s) = s$. A linear forward SDE
\begin{equation}
  \mathrm{d}x = f(x,s)\,\mathrm{d}s + g(s)\,\mathrm{d}w, \qquad f(x,s) = h(s)\,x,
  \label{eq:sde}
\end{equation}
started at $x_1 \sim p_{\rm data}$ has mean and variance functions as
\begin{equation}
  \dot{\alpha} = h(s)\,\alpha(s), \qquad
  \dot{\sigma^2} = 2\,h(s)\,\sigma^2(s) + g^2(s).
  \label{eq:moments}
\end{equation}

Substituting $\alpha(s) = 1-s$ into the first equation gives $h(s) = -1/(1-s)$,
and then substituting $\sigma^2(s) = s^2$ into the second gives
\begin{equation}
  g^2(s) \;=\; 2s + \frac{2s^2}{1-s} \;=\; \frac{2s}{1-s}.
\end{equation}
 Therefore, we have
\begin{equation}
   f(x,s) = -\frac{x}{1-s}, \quad g^2(s) = \frac{2s}{1-s},
  \label{eq:fg}
\end{equation}
Using the flow transport time  $t$, which reaches
\begin{equation}
    f(x, 1-t) = -\frac{x}{t}, \quad g(1-t) = \sqrt{\frac{2(1-t)}{t}}.
\end{equation}

% \subsection{The velocity is a gradient field with Gaussian linear interpolant}

% The marginal velocity of a Gaussian probability path satisfies
% $v_t(x) = (\dot\alpha_t/\alpha_t)\,x + b_t \nabla_x \log p_t(x)$ with
% $b_t = \sigma_t^2 \dot\alpha_t/\alpha_t - \sigma_t \dot\sigma_t$, and is therefore the
% gradient of
% \begin{equation}
%   \Phi(x,t) \;=\; \frac{\dot\alpha_t}{2\alpha_t}\lVert x\rVert^2 \;+\; b_t \log p_t(x).
%   \label{eq:remark}
% \end{equation}
% Parameterizing the velocity as $\nabla_x \Phi_\theta$ thus imposes no restriction at the
% optimum of the flow-matching objective \cite{lipman2023fm}, while making the marginal energy an explicit
% algebraic function of the learned potential. The construction requires the conditional
% to be Gaussian with isotropic covariance; for non-Gaussian sources or anisotropic noise
% the marginal velocity need not be curl-free. For our path
% ($\alpha_t = t$, $\sigma_t = 1-t$) one has $\dot\alpha_t/\alpha_t = 1/t$ and
% $b_t = (1-t)^2/t + (1-t) = (1-t)/t$, so \eqref{eq:remark} reduces to
% $\Phi = \lVert x \rVert^2/(2t) + \frac{1-t}{t}\log p_t$, consistent with
% \eqref{eq:v_from_score}.

\subsection{Terminal Energy}
By setting the constant $c(t)\equiv 0$, we obtain the energy of transport marginals as follows,
\begin{equation}
  \;E_t(x,t) \;=\; \frac{ \lVert x\rVert^2 - 2t\,\Phi(x,t)}{2(1-t)}\;
  \label{eq:Et}
\end{equation}
We use this quantity faithfully for $t \in[0, t_{\max}]$. For \emph{qualitative} visualization in  Figure 1 and AUROC computation,  we use $t_{\max} = 0.99999$ and  a simplified scalar energy as $E(x) = \Vert x \Vert^2 - 2 \Phi(x, t_{\max})$, dropping the positive constant $1/(1-t_{\max})$ and the factor $t_{\max}$, which largely preserves the energy landscape geometry and has minimal impact on AUROC values.  For EnergyPC usage in unconditional generation and inverse problem experiments, we set the noise floor $\sigma_{\min} = 10^{-3}$, hence the corresponding terminal time $t_{\max} = 0.999$.

%======================================================================
\section{Potential flow model training details}
%======================================================================
We implement the framework in PyTorch 2.11 (CUDA 12.8, Python 3.12) and run all training and evaluation on a single NVIDIA A100-SXM4-40GB GPU; the Python environment is managed by \texttt{uv} and pinned by the \texttt{uv.lock} file shipped with the code.

\paragraph{Scalar-potential parameterization.}
The network is a scalar-valued map $\Phi_\theta: \mathbb R^{d} \times [0,1] \to \mathbb R$
formed from the \emph{dot product} between an ADM/guided-diffusion UNet backbone $N_\theta$ \cite{dhariwal2021diffusionmodelsbeatgans} and the input itself $x$:
\begin{equation}
  \Phi_\theta(x,t) \;=\; \big\langle x,\; N_\theta(x,t)\big\rangle,
\end{equation}
with $\nabla_x\Phi_\theta$ computed by automatic differentiation and being used in the training objective. The dot product formulation has, in both our experiment and a previous work \cite{Du2023-ms}, had better empirical results than the denoising autoencoder objective form in \cite{Salimans2021-ok}. Time conditioning enters as two extra scalars appended to the flattened state, $t$ and $\sigma = 1-t$; $t$ is passed through the usual sinusoidal timestep embedding.

\paragraph{Data normalization.}
The data is standardized \emph{channel-wise} based on \emph{fixed} values ported from DiffuisonPDE codebase during both training and sampling.
\begin{equation}
    x^{\rm norm}_c = \frac{x_c - \mu_c }{\sigma_c}
\end{equation}
While distributional metrics are computed in the normalized space, we use the same statistics to rescale the generated samples back to the original physical scale to the compute PDE residuals.
\begin{equation}
    x_c = \sigma_cx^{\rm norm}_c + \mu_c
\end{equation}

\paragraph{Architecture.}
We use the same model architecture and hyperparameters for the three datasets, Poisson, Helmholtz, and Burgers. The detailed hyperparameter configuration are listed in the following table.
\begin{center}
\small
\begin{tabular}{lll}
\toprule
Hyperparameter & Value & Note \\
\midrule
\multicolumn{3}{l}{\emph{Backbone}}\\
\texttt{dims}                  & $2$                & for 2D PDE fields\\
\texttt{data.image\_size}      & $128$              & fields are $128\times128$ \\
\texttt{data.channels}   & $2$ / $1$          & Poisson, Helmholtz / Burgers \\
\texttt{nf}                    & $128$              & base width \\
\texttt{channel\_mult}         & $[1,2,2,4]$        & widths $128,256,256,512$ at $128^2,64^2,32^2,16^2$ \\
\texttt{num\_res\_blocks}      & $4$                & per resolution \\
\texttt{activation}                  & \texttt{silu}      & \\
\texttt{use\_scale\_shift\_norm} & \texttt{true}    & FiLM-style conditioning \\
\texttt{conv\_resample}        & \texttt{false}     & no learned resampling \\
\texttt{resblock\_updown}      & \texttt{false}     & \\
\midrule
\multicolumn{3}{l}{\emph{Attention}}\\
\texttt{attention\_resolutions} & $[4, 8]$          & downsample factors, i.e.\ $32^2$ and $16^2$ \\
\texttt{num\_head\_channels}   & $64$               & sets head count; $19$ blocks total \\
heads (derived)                & $4$ / $8$          & $256/64$ at $32^2$, $512/64$ at $16^2$ \\
\texttt{use\_new\_attention\_order} & \texttt{true} & \\
\midrule
\multicolumn{3}{l}{\emph{Conditioning, regularization, head}}\\
\texttt{temb\_type}            & \texttt{time}      & sinusoidal embedding of $t$; $\sigma=1-t$ appended \\
\texttt{dropout}               & $0.13$             & \\
\texttt{with\_fourier\_features} & \texttt{false}   & \\
\texttt{num\_classes}          & \texttt{null}      & unconditional; no CFG \\
output head                    & $\langle x, N_\theta(x,t)\rangle$ & zero-init $3\times3$ conv, then contract \\
\midrule
Parameter count                    & 121M  &  \\
\bottomrule
\end{tabular}
\end{center}

\paragraph{Model training.}
Table below lists the details of the implemented training objective, optimizer parameters, and number of gradient steps used for model training.  The reported evaluation results use the raw (non-EMA) weights at step $34000$ for Poisson and at the final step $40000$ for Helmholtz and Burgers.
\begin{center}
\small
\begin{tabular}{lll}
\toprule
Hyperparameter & Value & Note \\
\midrule
\multicolumn{3}{l}{\emph{Interpolant and time sampling}}\\
\texttt{scheduling} / \texttt{fm\_target} & \texttt{fmot}    & conditional-OT path; target $v^\star = x_1 - \varepsilon$ \\
\texttt{t\_start}, \texttt{t\_max}        & $10^{-5}$, $0.99999$ & $t \sim \mathcal U[t_{\rm start}, t_{\max}]$ per step \\
\texttt{mean\_power}, \texttt{var\_power} & $1$, $1$         & $\alpha_t = t$, $\sigma_t = 1-t$ \\
\texttt{sigma\_eps}                       & $0$              & no additive noise floor on $\sigma_t$ \\
\texttt{augment\_t}                       & \texttt{true}    & $t$ and $\sigma_t$ appended to the flat input \\
\texttt{shift\_schedule}                  & \texttt{false}   & no resolution-dependent $t$ shift \\
\midrule
\multicolumn{3}{l}{\emph{Optimizer}}\\
\texttt{optimizer}                        & \texttt{Adam}    & \\
\texttt{lr}                               & $2\times10^{-4}$ & \\
\texttt{beta1}                            & $0.9$            & $\beta_2 = 0.999$ derived, not configured \\
\texttt{eps}                              & $10^{-8}$        & \\
\texttt{weight\_decay}                    & $0$              & \\
\texttt{warmup}                           & $5000$           & linear \\
\texttt{anneal\_rate}, \texttt{anneal\_iters} & $1$, $[0,0]$ & no LR decay \\
\texttt{grad\_clip}, \texttt{grad\_clip\_mode} & $3$, \texttt{std} & clip to $3\sqrt{\hat m_2} + 0.1$ per parameter \\
\texttt{batch\_size}                      & $4$              & \texttt{small\_batch\_size} $=4$, so no subsetting \\
\texttt{n\_iters}                         & $40{,}000$       & \\
\texttt{seed}                             & $49$             & \\
% \multicolumn{3}{l}{\emph{EMA}}\\
% \texttt{ema\_rate}                        & $0.9999$         & tracked; stored only in \texttt{last\_full.pt} \\
% reported weights                          & non-EMA          & \texttt{best.pt}/\texttt{last.pt} are raw \texttt{state\_dict} \\
\bottomrule
\end{tabular}
\end{center}

%======================================================================
\section{Additional details on EnergyPC}
\label{sec:samplers}
%======================================================================
All samplers act on the normalized space. Samples are rescaled by to the original physical space for PDE residual computation and visualization. 
\paragraph{Noise ladder and predictor.}
  The network is conditioned on $t$, so the ladder is built in the noise scale
  $\sigma(t) = 1-t$ and converted back. We place the given number of levels, $L$, geometrically in
  $\sigma$,
  \begin{equation}
  \sigma_i = \sigma_{\max}\Big(\tfrac{\sigma_{\min}}{\sigma_{\max}}\Big)^{i/(L-1)},
  \qquad t_i = 1-\sigma_i,
  \end{equation}
  with $\sigma_{\max} = 1-t_{\min} = 1-10^{-5}$ and $\sigma_{\min} =  10^{-3}$, so $t$ runs from $10^{-5}$ to $0.999$. The predictor is one explicit Euler step of the probability-flow ODE, $x \leftarrow x + \sigma_{\min} \,\nabla_x \Phi \theta(x,1-\sigma_{\min})$, after which the corrector runs at $t_{i+1}$.

\paragraph{EnergyPC target.} 
The EnergyPC targets the physics-augmented total energy
\begin{equation}
  E_{\rm tot}(x) = E_t(x) \;+\; \lambda_t\rVert R\rVert^2, \qquad
  R(\xhat) = \frac{F(\xhat)}{N} ,
  \label{eq:U}
\end{equation}
with $E_t$ from \eqref{eq:Et}, $N$ is the number of discrete points for the length of the domain, and $\mathcal F$ the discrete PDE residual operator of Section~\ref{sec:data}.  We use the leaf gradient, $\nabla_{\xhat} \lambda_t\Vert R\Vert^2$, of the physical energy term to get Langevin proposals;  Metropolis steps direclty use $E_{\rm tot}$, so the sampler targets $e^{-E_{\rm tot}}/Z$.

 The step size is chosen per sample and per level as
  \begin{equation}
    \eta = \min\!\Big(2\big(\mathrm{snr}\,\sqrt{d}\,/\,\lVert\hat s\rVert\big)^2,\;
    c\,\sigma^2\Big), \qquad \mathrm{snr} = 0.1, \quad c = 0.5,
  \end{equation}
  where $d$ is the flattened state dimension and $\hat s = -\frac12\nabla E_{\rm tot}$ is the corrector drift of Eq.~\eqref{eq:U}. The first term follows the SNR-adaptive rule of \cite{Song2020-iv}, sizing the move relative to the drift norm, with two changes: we evaluate it per sample rather than on the batch-averaged norms, and use $\sqrt{d} \approx \mathbb{E}\lVert z\rVert$ in place of the sampled noise norm. The second term is the classic NCSN step \cite{Song2019-vb}, acting as a cap. The adaptive step is what allows the physics term to be run unclipped. Where $\hat x_1$ is still off-manifold the residual gradient is stiff and $\lVert\hat s\rVert$ is large, and the rule shrinks $\eta$ there automatically; a fixed $c\sigma^2$ step would instead require clipping the drift, which would cost MALA its exact target. Evaluating it per sample matters for the same reason, since a single off-manifold sample is what diverges and a batch-averaged step would size the move from the others. The rule also keeps acceptance nonzero at the top of the ladder, where a fixed step overshoots and is rejected almost surely, while the $c\sigma^2$ cap bounds the move at the other end, when $\lVert\hat s\rVert$ is small.
  
\paragraph{Physics weight schedule $\lambda_t$.}
In the reported PC rows the physics weight is gated on the noise scale,
  \begin{equation}
    \lambda_i = \lambda_{\max}\cdot
    \begin{cases}
      0, & \sigma_i \ge 0.3,\\[2pt]
      \dfrac{0.3 - \sigma_i}{0.25}, & 0.05 < \sigma_i < 0.3,\\[6pt]
      1, & \sigma_i \le 0.05,
    \end{cases}
    \qquad \lambda_{\max} = 2 .
  \end{equation}
  At high noise $\xhat$ is far off-manifold, so $\lVert R\rVert^2$ and its gradient are
  large and dominate $\lVert \hat s\rVert$; the gate confines physics to the band where
  the Tweedie estimate is meaningful. Since the schedule multiplies $\lambda_{\max}$, it
  is identically zero when $\lambda_{\max} = 0$: the ``Standard PC'' rows run the same
  code path with the physics term absent at every level.

\paragraph{Inverse problems: EnergyPC on $E_{\rm pos}$}
For the inverse experiments the predictor is an unmodified faithful DiffusionPDE or
FlowDPS guidance step and the corrector targets the true posterior energy
\begin{equation}
  E_{\rm pos}(x,t) \;=\; E_{\rm tot}(x,t)
  \;+\; \frac{1}{2\sigma_{\rm obs}^2}\big\lVert m \odot (\xhat - o)\big\rVert_2^2,
  \label{eq:Upost}
\end{equation}
where $m$ is the $0/1$ observation mask, $o$ the normalized observations, and
$\sigma_o = \texttt{noise}/\texttt{std}_{\rm channel}$ (values listed below) the true observation-noise
standard deviation in normalized space. The observation term is the squared residual, so \eqref{eq:Upost} is a proper Gaussian negative log-likelihood and the MH ratio is exact. 
\begin{center}
  \begin{tabular}{lccc}
  \toprule
  Dataset & Noise (physical) & $\texttt{std}_{\rm channel}$ & $\sigma_o$ \\
  \midrule
  Poisson   & $5.5{\times}10^{-4}$ & $1/36.5 = 0.027397$ & $0.0201$ \\
  Helmholtz & $5.6{\times}10^{-4}$ & $0.028$             & $0.0200$ \\
  \bottomrule
  \end{tabular}
  \end{center}

\paragraph{Predictor settings.}
Both baselines are used unmodified at their released defaults, shown in the table below.
Both guide the Tweedie estimate $\xhat = x + \sigma\,\nabla_x \Phi_\theta(x,t)$ with $L_2$-\emph{norm}
losses. Note that FlowDPS \cite{Kim2025-sa} is
observation-only,  whereas DiffusionPDE
\cite{Huang2024-rv} guides on both the observations and the residual.
\begin{center}
\begin{tabular}{lll}
\toprule
Method & Parameter & Value \\
\midrule
DiffusionPDE & steps & $200$ \\
& $\zeta_{\rm obs}$, $\zeta_{\rm pde}$ & $1$, $1$ \\
& PDE guidance start & after $80\%$ of steps \\
& late obs.\ scale & $0.1$ \\
& guidance-norm clip & $50$ \\
\midrule
FlowDPS & steps & $200$ \\
& DC step size & $30.0$ \\
& DC inner iterations & $3$ \\
& physics guidance & none \\
\bottomrule
\end{tabular}
\end{center}

%======================================================================
\section{PDE datasets}
\label{sec:data}
%======================================================================
All three datasets are the DiffusionPDE release \cite{Huang2024-rv}
(\url{https://github.com/jhhuangchloe/DiffusionPDE}),
read from the distributed \texttt{.mat} files at $128 \times 128$ resolution. Each has
$10000$ samples split contiguously into $9000$ train (indices $[0,9000)$)
and $1000$ test (indices $[9000,10000)$); every evaluation in the paper draws from the
test split only.

\begin{center}
\begin{tabular}{llll}
\toprule
Dataset & Equation & Channels & Residual operator \\
\midrule
Poisson & $\Delta u = a$ on $(0,1)^2$, $u|_{\partial\Omega} = 0$
        & $[a, u]$, $C{=}2$ & $\mathcal F = \Delta_h u - a$ \\
Helmholtz & $\Delta u + u = a$ on $(0,1)^2$, $u|_{\partial\Omega} = 0$
        & $[a, u]$, $C{=}2$ & $\mathcal F = \Delta_h u + u - a$ \\
Burgers & $u_t + u u_x - \nu u_{xx} = 0$, $\nu = 0.01$
        & $[u(t,x)]$, $C{=}1$ & $\mathcal F = u_t + u u_x - \nu u_{xx}$ \\
\bottomrule
\end{tabular}
\end{center}

\paragraph{Discretization of the residual operators.}
PDE residual calculations are directly ported from the DiffusionPDE codebase, matching their discretization exactly. Poisson and Helmholtz use the $5$-point Laplacian with zero padding and grid spacing $h = 1/(S-1)$, $S = 128$, and the four boundary edges are zeroed after applying the stencil. Burgers uses central-difference $3$-tap stencils in both axes with unit grid spacing ($\mathrm{d}t = \mathrm{d}x = 1$) and zero padding, with rows indexing time and columns space on a $128 \times 128$ space--time grid. 

\paragraph{Channel-wise statistics.}
Following DiffusionPDE, fields are normalized by a \emph{fixed} per-channel standardization, where the mean is $0$ for every channel, and standard deviation listed as follows

\begin{center}
\begin{tabular}{llrr}
\toprule
Dataset & Channel & mean $\mu_c$ & std $s_c$ \\
\midrule
\multirow{2}{*}{Poisson}   & $f$ (source)   & $0$ & $2.15$ \\
                         & $\phi$ (solution) & $0$ & $0.027397$ \\
\midrule
\multirow{2}{*}{Helmholtz} & $a$ (forcing)  & $0$ & $2.15$ \\
                         & $u$ (wave)     & $0$ & $0.028$ \\
\midrule
Burgers                    & $u$            & $0$ & $1.415$ \\
\bottomrule
\end{tabular}
\end{center}

%======================================================================
\section{Evaluation Metrics}
\label{sec:metrics}
%======================================================================
Distributional metrics (sliced-$W_2$, spectral distance, MMSE/SMSE) are computed in the \emph{normalized coordinates}, which weights channels of very different physical scale comparably. The PDE residual is computed in the \emph{original} physical scale. 
  \paragraph{Sliced $W_2$.}
  Given the sample size limit ($m=100$), we use sliced Wasserstein distance \cite{kolouri2019generalizedslicedwassersteindistances} to measure the distributional discrepancy using the generated samples and true data samples
\begin{equation}
    \mathrm{SW}_2(\widehat X, X)
    = \Big(\mathbb E_{\omega}\big[W_2^2(\omega_{\#}\widehat X,\, \omega_{\#}X)\big]\Big)^{1/2},
  \end{equation}
  where $\omega$ is a random unit vector in $\mathbb R^{d}$, $\omega_{\#}$ denotes the
  distribution of the projections $\omega^\top x$, 
  $X = \{x^{(i)}\}_{i=1}^{n}$ denotes the real fields (the test split), and $\widehat X = \{\hat x^{(j)}\}_{j=1}^{m}$ the generated ensemble.  
  The sliced-$W_2$ is estimated with $256$ directions drawn i.i.d.\ Gaussian and normalized to the unit sphere, and $128$ evenly spaced quantiles per projection. 

  \paragraph{Radial log-spectrum $L_1$.}
  We compare generated and reference fields through their radially averaged power spectra, the standard spectral-fidelity check for generative field models \cite{durall2020watchupconvolutioncnnbased, lienen2024zeroturbulencegenerativemodeling}.  For each channel $c$ we take the squared magnitude of the 2D DFT of every sample and average it over the $m$ samples, giving one power value per frequency mode $k$. We then reduce this to a 1D profile by averaging over direction, where all modes whose distance from the zero-frequency center rounds to the same integer $r$ are averaged together,
  \begin{equation}
    \widehat E_c(r) = \frac{1}{\lvert A_r\rvert}\sum_{k \in A_r}
    \frac{1}{m}\sum_{j=1}^{m}\big\lvert \mathcal F \hat x^{(j)}_c(k)\big\rvert^2 ,
    \qquad A_r = \{k : \mathrm{round}(\lVert k\rVert) = r\},
  \end{equation}
  so that $r$ indexes spatial frequency, in wavelengths across the domain. On the
  $128\times128$ grid this gives $r = 0,\dots,R$ with $R = 91$, the radius of the corner mode. Similarly, we compute $E_c(r)$ from the $n$ real test fields and report the
  mean \emph{absolute} difference of the log profiles over bins and channels,
  \begin{equation}
    \mathrm{Spectral}(\widehat X, X)
    = \frac{1}{C(R+1)}\sum_{c=1}^{C}\sum_{r=0}^{R}
      \big\lvert \log_{10}\widehat E_c(r) - \log_{10} E_c(r)\big\rvert .
  \end{equation}
  This is the log-spectrum distance of \cite{lienen2024zeroturbulencegenerativemodeling} with $L_1$ in place
  of $L_2$ and a per-bin mean in place of the norm, so the reported value is the mean absolute $\log_{10}$ discrepancy per radial bin. This measures the discrepancy of energy at each spatial scale between the generated samples and real data.

  \paragraph{PDE residual.}
  The reported column is the ensemble mean of $\lVert R(\hat x)\rVert^2$, with
  $R = \mathcal F(\hat x)/N$ as in \eqref{eq:U}.

  \paragraph{MMSE / SMSE.}
  Let $\bar{\hat x}, s_{\hat x}$ be the per-pixel mean and standard-deviation
  fields over the $m$ generated samples, and $\bar x, s_x$ the same over the $n$ test
  fields. Then
  \begin{equation}
    \mathrm{MMSE} = \frac{\lVert \bar{\hat x} - \bar x\rVert^2}{CN^2},
    \qquad
    \mathrm{SMSE} = \frac{\lVert s_{\hat x} - s_x\rVert^2}{CN^2},
  \end{equation}
  where $C$ is the number of channels in the data.
  MMSE detects mean/bias error, SMSE detects spread mismatch.

% \paragraph{Reference floors.}
% Every distributional metric is reported against a data-vs-data floor: two disjoint
% halves of the test reference for sliced $W_2$ and spectral, and for MMSE/SMSE an
% $n$-sample data subset scored against the disjoint remainder, averaged over $8$ random
% splits.
% \begin{center}
% \begin{tabular}{lcccc}
% \toprule
% Dataset & Sliced-$W_2$ & Spectral & MMSE & SMSE  \\
% \midrule
% Poisson   & $0.0428$ & $0.0936$ & $2.81{\times}10^{-4}$ & $1.21{\times}10^{-4}$  \\
% Helmholtz & $0.0415$ & $0.0647$ & $1.59{\times}10^{-4}$ & $1.36{\times}10^{-4}$  \\
% Burgers   & $0.0441$ & $0.0246$ & $2.04{\times}10^{-4}$ & $0.78{\times}10^{-4}$  \\
% \bottomrule
% \end{tabular}
% \end{center}
% The MMSE/SMSE floors are Monte-Carlo estimates over $8$ random splits and carry their
% own error: two Burgers runs of the same configuration recorded $2.04{\times}10^{-4}$ and
% $2.66{\times}10^{-4}$ (MMSE), so treat them as an order-of-magnitude reference rather
% than a sharp threshold. An exact empirical $W_2$ via optimal assignment is also stored
% in every JSON but is \emph{not} reported: at $n \approx 100$ its data-vs-data floor
% exceeds every method's value, so it carries no information.

\paragraph{OOD scores and AUROC.}
Two terms are computed independently of any weight: the learned prior read-off
$E = \lVert x\rVert^2 - 2\Phi(x,t_{\max})$ and the mean-square residual $\Rres$. Their
scales differ by orders of magnitude, so the reported total is scale-balanced,
\begin{equation}
  E_{\rm tot} = E + 2\lambda_{\rm bal} \Rres, \qquad
  \lambda_{\rm bal} = \frac{\mathrm{std}(E)}{2\,\mathrm{std}(\Rres)},
\end{equation}
so that both terms carry the same in-distribution spread and $E_{\rm tot}$ flags a field if \emph{either} term does. The fitted values are $\lambda_{\rm bal} = 1.729{\times}10^{11}$ (Poisson), $1.018{\times}10^{11}$ (Helmholtz), and $5.412{\times}10^{4}$ (Burgers). 

\paragraph{OOD corruption tiers.}
We construct the out-of-distribution sets as follows. \emph{Gaussian} draws $\mathcal N(0,I)$ directly in normalized space. For \emph{Noise $s$} we add Gaussian noise with standard deviation $s \in \{0.1, 0.5\}$ times each channel's own standard deviation, so the corruption is field-relative rather than absolute. For \emph{Blur} we apply a separable Gaussian smoothing (kernel size $9$, $\sigma = 3.0$, reflect padding) to all channels with the same kernel;  For \emph{Roll} we cyclically shift the field by $H/4$ along both axes, and for \emph{Channel shuffle} we pair $a_i$ with $u_j$, $i \ne j$, which applies to the two-channel datasets only (Poisson and Helmholtz). Finally, \emph{Cross-PDE} uses the test fields of the other dataset (Poisson$\leftrightarrow$Helmholtz).

\paragraph{Inverse-problem error.}
For a field $F \in \{a, u\}$ with ground truth $F^\star$ and ensemble mean $\bar F$ over
$M = 16$ draws, in physical units,
\begin{equation}
  \mathrm{RE}(F) = \frac{\lVert \bar F - F^\star\rVert}{\lVert F^\star \rVert} ,
\end{equation}
reported as aRE for the recovered channel $a$ and uRE for the observed channel $u$.
Observations are $500$ pixels of channel $u$ drawn uniformly without replacement (fixed
across methods and problems), corrupted by Gaussian noise with standard deviation
$5.5{\times}10^{-4}$ (Poisson) or $5.6{\times}10^{-4}$ (Helmholtz), $\approx 2\%$ of the
field standard deviation.

\paragraph{Confidence intervals.}
All confidence intervals are computed from bootstrapped $2000$ samples. The
reference data is held fixed so the interval reflects generator variability
only. For the inverse table the $16$ posterior draws are resampled and the
ensemble mean re-formed ($5000$ resamples).

\section{Additional results}
Access to the energy functions enables pure MCMC sampling either from the ODE samples or  Gaussians. We include the results of two additional samplers in this section. 

\paragraph{Energy-based refinement of ODE samples.}
We run MCMC refinement using the terminal time energy $E_{\rm tot}(x, 1-\sigma_{\min})$ with chains initialized by the baseline ODE samples. We run $100$ steps of ULA and MALA using the SNR-adaptive step size $\eta(x) = \min\big(2(\mathrm{snr}\sqrt D / \lVert \hat s(x)\rVert)^2, \, c\,\sigma_{\min}^2\big)$ with $\mathrm{snr} = 0.1$, $c = 0.5$ and $\sigma_{\min} = 10^{-3}$.
\paragraph{Energy annealing MCMC}
Independent of the ODE sampler, we use the time-dependent (noise-level dependent) energy functions to run annealed MCMC to generate samples from Gaussians. Starting from $x \sim \mathcal N(0, I)$, we sweep the same geometric ladder of $120$ noise levels used by the ODE sampler, $\sigma = 1 \to 10^{-3}$, and take $K = 8$ Langevin steps at each level with the fixed step size $\delta = c\,\sigma^2$, $c = 0.5$, followed by a single Tweedie denoise. The physics weight is gated by noise level: $\lambda = 0$ for $\sigma \ge 0.3$, ramping linearly to $\lambda_{\max}$ as $\sigma$ falls to $0.05$ and held there below. The steps are ULA for $\sigma \ge 0.3$ and switched to MALA when $\sigma < 0.3$. 
\begin{figure}
    \centering
    \includegraphics[width=\linewidth]{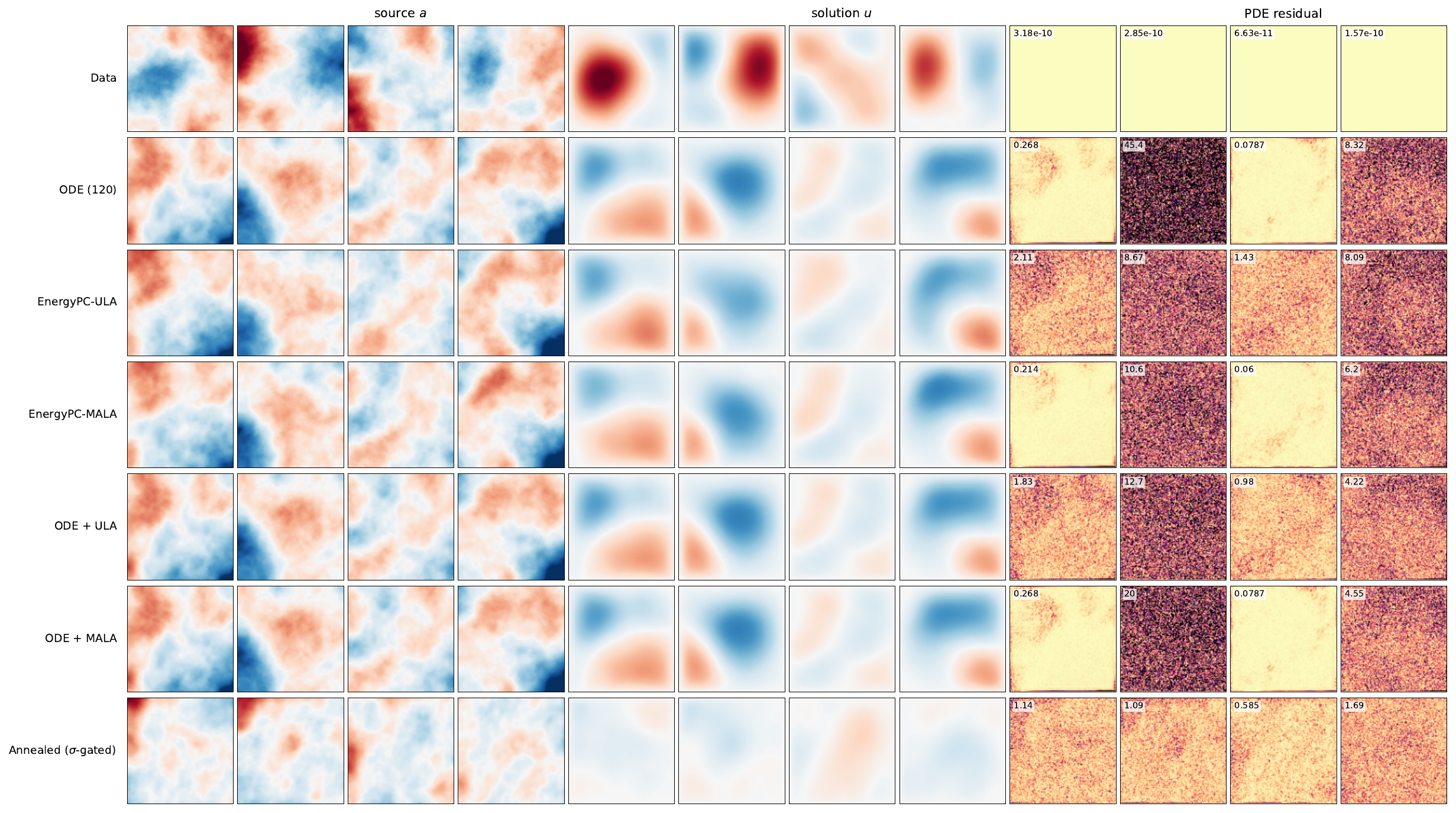}
    \caption{Unconditional generation of Poisson equation data with various samplers. All energy-based samplers reduce the PDE residuals compared to the baseline flow ODE sampler.}
    \label{fig:sampler_grid_poisson}
\end{figure}

\begin{table}[t]\centering\small
  \caption{Additional unconditional samplers on Poisson, not reported in the main text. }
  \label{tab:extra_samplers_poisson}
  \begin{tabular}{llrcc}\toprule
  Sampler & physics & NFE & Sliced-$W_2\!\downarrow$ & PDE resid$\downarrow$ \\\midrule
  ODE120$^\dagger$ & -- & 120 & 0.0532 & 13.66 \\
  ODE + ULA polish & terminal & 220 & 0.0408 & 10.33 \\
  ODE + MALA polish & terminal & 320 & 0.0409 & 7.17 \\
  Annealed + $\sigma$-sched & gated & 1752 & 0.0880 & 1.15 \\
  \bottomrule\end{tabular}\end{table}

\begin{figure}
    \centering
    \includegraphics[width=\linewidth]{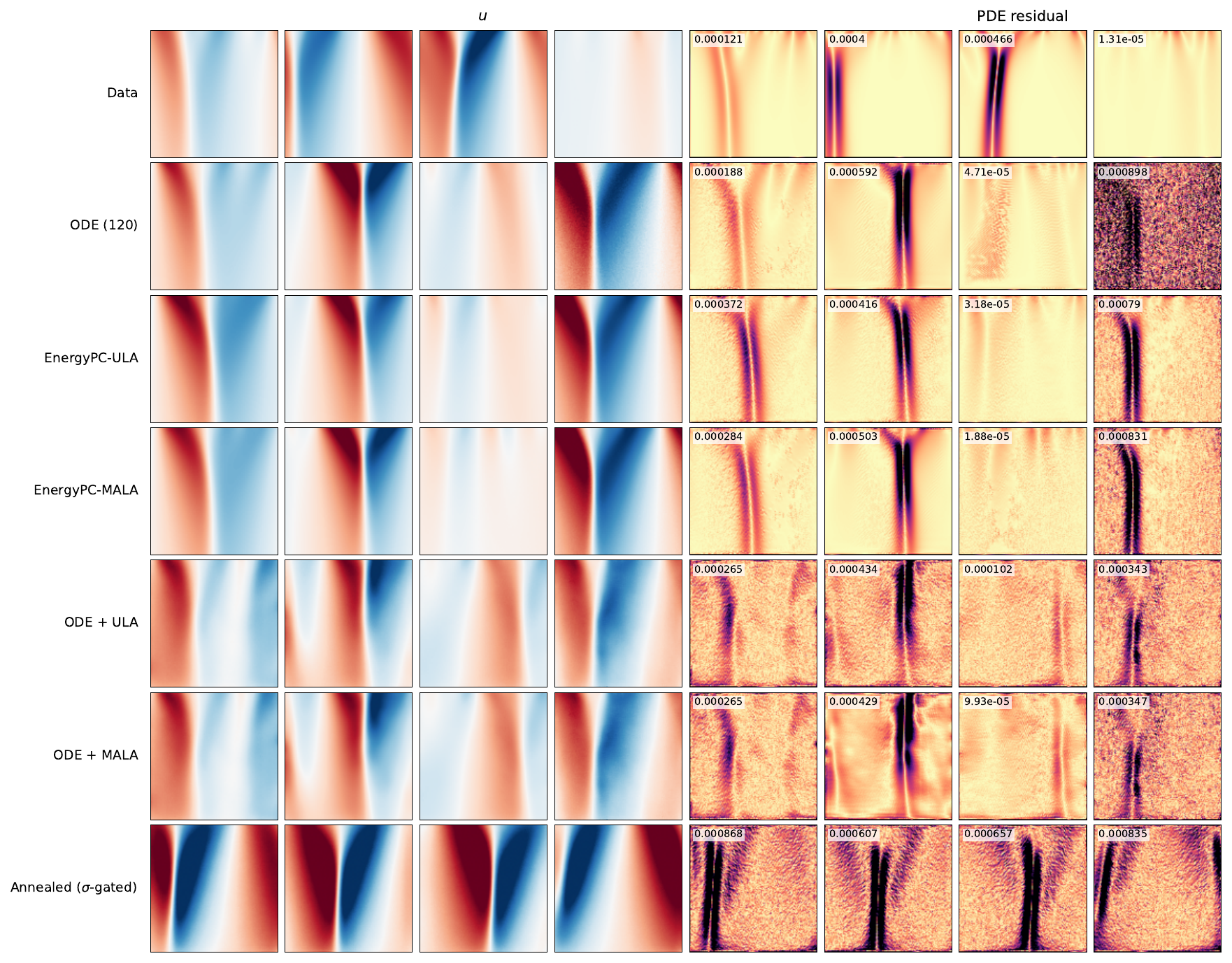}
    \caption{Unconditional generation of Burgers equation data with various samplers. Energy-based samplers, except the energy annealing sampler, reduce the PDE residual compared to the ODE baseline sampler. }
    \label{fig:sampler_grid_burgers}
\end{figure}

  \begin{table}[t]\centering\small
  \caption{Additional unconditional samplers on Burgers, not reported in the main text.}
  \label{tab:extra_samplers_burgers}
  \begin{tabular}{llrcc}\toprule
  Sampler & physics & NFE & Sliced-$W_2\!\downarrow$ & PDE resid$\downarrow$ \\\midrule
  ODE120$^\dagger$ & -- & 120 & 0.0521 & 3.59 \\
  ODE + ULA polish & terminal & 220 & 0.0531 & 2.58 \\
  ODE + MALA polish & terminal & 320 & 0.0513 & 2.54 \\
  Annealed + $\sigma$-sched & gated & 1752 & 0.1928 & 7.73 \\
  \bottomrule\end{tabular}\end{table}

% \bibliography{aaai2027, paperpile, ref}

% \end{document}

\end{document}